\documentclass[acmsmall]{acmart}

\AtBeginDocument{%
  \providecommand\BibTeX{{%
    \normalfont B\kern-0.5em{\scshape i\kern-0.25em b}\kern-0.8em\TeX}}}

\setcopyright{acmcopyright}
\copyrightyear{2022}
\acmYear{2022}
\acmDOI{XXXXXXX.XXXXXXX}

\acmBooktitle{Woodstock '18: ACM Symposium on Neural Gaze Detection,
 June 03--05, 2018, Woodstock, NY} 
\acmPrice{15.00}
\acmISBN{978-1-4503-XXXX-X/18/06}

\usepackage{multirow}
\usepackage{subfigure}
\usepackage[inline]{enumitem}
\DeclareUnicodeCharacter{2212}{-}

\begin{document}

\title{Lightweight Deep Learning for Resource-Constrained Environments: A Survey}

\author{Hou-I Liu}
\email{k39967.c@nycu.edu.tw}
\orcid{0000-0002-2101-2997}
\affiliation{%
  \institution{Department of Electronics and Electrical Engineering, National Yang Ming Chiao Tung University}
  \city{Hsinchu}
  \country{Taiwan, ROC}
  \postcode{300}
}

\author{Marco Galindo}
\email{marcodavidg@gmail.com}
\affiliation{%
{ \institution{Department of Electrical Engineering and Computer Science, National Yang Ming Chiao Tung University}}
  \city{Hsinchu}
  \country{Taiwan, ROC}
  \postcode{300}
}

\author{Hongxia Xie}
\affiliation{%
  \institution{College of Computer Science and Technology, Jilin University}
  \institution{Key Laboratory of Symbolic Computation and Knowledge Engineering of Ministry of Education, Jilin University}
  \city{Changchun}
\country{China}
  \postcode{130000}
}
\email{hongxiaxie.ee08@nycu.edu.tw}

\author{Lai-Kuan Wong}
\email{lkwong@mmu.edu.my}
\affiliation{%
  \institution{Faculty of Computing and Informatics, Multimedia University}
  \city{Cyberjaya}
  \country{Malaysia}
  \postcode{63100}
}

\author{Hong-Han Shuai}
\affiliation{%
  \institution{Department of Electronics and Electrical Engineering, National Yang Ming Chiao Tung University}
  \city{Hsinchu}
\country{Taiwan, ROC}
  \postcode{300}
}
\email{hhshuai@nycu.edu.tw}

\author{Yung-Hui Li}
\affiliation{%
  \institution{Hon Hai Research Institute}
  \city{Taipei}
\country{Taiwan, ROC}
  \postcode{114}
}
\email{yunghui.li@foxconn.com}

\author{Wen-Huang Cheng}
\affiliation{%
  \institution{Department of Computer Science and Information Engineering, National Taiwan University}
  \city{Taipei}
\country{Taiwan, ROC}
  \postcode{106}
}
\email{wenhuang@csie.ntu.edu.tw}

\renewcommand{\shortauthors}{H.-I. Liu et al.}

\begin{abstract}
Over the past decade, the dominance of deep learning has prevailed across various domains of artificial intelligence, including natural language processing, computer vision, and biomedical signal processing. While there have been remarkable improvements in model accuracy, deploying these models on lightweight devices, such as mobile phones and microcontrollers, is constrained by limited resources. In this survey, we provide comprehensive design guidance tailored for these devices, detailing the meticulous design of lightweight models, compression methods, and hardware acceleration strategies. The principal goal of this work is to explore methods and concepts for getting around hardware constraints without compromising the model's accuracy. Additionally, we explore two notable paths for lightweight deep learning in the future: deployment techniques for TinyML and Large Language Models. Although these paths undoubtedly have potential, they also present significant challenges, encouraging research into unexplored areas.
\end{abstract}

\ccsdesc[500]{Computing methodologies~Neural networks}
\ccsdesc[500]{Computing methodologies~Artificial intelligence}
\ccsdesc[500]{Computing methodologies~Computer vision}
\ccsdesc[500]{Computing methodologies~Model compression}
\ccsdesc[500]{Computer systems organization~Embedded systems}
\ccsdesc[500]{Software and its engineering~Designing software}
\ccsdesc[500]{Software and its engineering~Software design techniques}

\keywords{Lightweight model, efficient transformer, model compression, quantization, tinyML, large language models}

\maketitle

\section{Introduction}
Over recent years, the importance of neural networks (NNs) has escalated tremendously, with their applications permeating various aspects of daily life and extending to support complex tasks~\cite{hidayati2020dress,xie2023most,chen2021fashionmirror}. However, since the publication of AlexNet~\cite{krizhevsky2012imagenet} in 2012, there has been a prevailing trend toward creating deeper and more intricate networks to enhance accuracy. For instance, Model Soups~\cite{wortsman2022model} has achieved remarkable accuracy on the ImageNet dataset, but it comes at the cost of over 1,843 million parameters. Similarly, GPT-4~\cite{gpt4-para} has demonstrated outstanding performance on natural language processing (NLP) benchmarks, albeit with a staggering 1.76 trillion parameters. Notably, Amodei et al.~\cite{AIComputeBlog} indicated that the computational demands of deep learning (DL) have surged dramatically, increasing by approximately 300,000 times from 2012 to 2018. This dramatic increase in size sets the stage for the challenges and developments explored in this paper.

Concurrently, Green AI~\cite{greenai,greenai2} has arisen as a prominent concern over the past few years, labeling hefty DL models unsuitable due to their substantial GPU and training time demands, which can contribute to environmental degradation. Strubell et al.~\cite{carbon} extensively analyze the carbon footprint of language models trained on multiple GPUs. In parallel, lightweight devices have garnered increased attention due to their versatile applications and portability. According to Sinha~\cite{Iphone}, the number of connected IoT devices grew by 18\% in 2022, reaching 14.4 billion, and has a projected escalation to 29.0 billion by 2027. A testament to this growing demand is the production of over 200 million iPhones since 2016. On the other hand, edge devices offer superior automation and energy efficiency compared to mobile devices, especially the deployment of ultra-low-cost microcontrollers (MCUs) in devices like pacemakers and forehead thermometers~\cite{theneedofmcu}. 

In response to the practical demands outlined above, a significant body of research has emerged in recent years, focusing on lightweight modeling, model compression, and acceleration techniques. The Annual Mobile AI (MAI) workshops have been held consecutively during CVPR 2021-2023~\cite{CVPRW1, CVPRW2, CVPRW3}, with a primary emphasis on the deployment of DL models for image processing on resource-constrained devices, such as ARM Mali GPUs and Raspberry Pi 4. Additionally, the Advances in Image Manipulation (AIM) workshops conducted at ICCV 2019, ICCV 2021, and ECCV 2022~\cite{AIM} have organized challenges centered around image/video manipulation, restoration, and enhancement on mobile devices. 

From our study, we discovered that the most effective approach for analyzing the development of an efficient, lightweight model, spanning from its design phase to deployment, involves incorporating three key elements into the pipeline: NN architecture design, compression methods, and hardware acceleration for lightweight DL models. 
Previous surveys~\cite{gou2021knowledge, Com2022, quan_2023, articleBerthelier, liang2021pruning} often focus on specific aspects of this pipeline, such as discussing only quantization methods, offering detailed insights into those segments. However, these surveys may not provide a comprehensive view of the entire process, potentially overlooking significant alternative approaches and techniques. In contrast, our survey covers lightweight architectures, compression methods, and hardware acceleration algorithms.

\subsection{Neural Network Design}
In the first part of this article, Section~\ref{section:Lightweight Architecture Design}, we examine the classic lightweight architectures, categorizing them into family series for improved clarity. Some of these architectures made significant strides by introducing innovative convolution blocks. For instance, depthwise separable convolutions~\cite{chollet2017xception} prioritize high accuracy and reduced computational demand. Sandler et al.~\cite{sandler2018mobilenetv2} introduce an inverted residual bottleneck to enhance gradient propagation. Other architectures, such as ShuffleNet~\cite{zhang2017shufflenet}, were able to develop an optimized convolution operation, which applies group convolution~\cite{krizhevsky2012imagenet} to achieve a parallel design and further improve the transferability between groups of data through shuffle operations. The ShiftNet~\cite{wu2018shift} achieves an equivalence effect of traditional convolution with no parameters or Floating Point Operations (FLOPs). The AdderNet~\cite{Chen_2020_CVPR} replaces the multiplication operation with the addition operation, greatly reducing computation requirements.

It is also important to note that parameters and FLOPs do not consistently correlate with inference time. Early lightweight architectures, such as SqueezeNet~\cite{iandola2016squeezenet} and MobileNet~\cite{howard2017mobilenets}, aim to reduce parameters and FLOPs. However, this reduction often increases Memory Access Cost (MAC)~\cite{ma2018shufflenet}, leading to slower inference. 
Hence, we aim to contribute to the application of lightweight models by providing a more comprehensive and insightful review.

\vspace{-0.5em}
\subsection{Neural Network Compression}
In addition to lightweight architecture designs, Section~\ref{section:Fundamental methods in model compression} mentions various efficient algorithms that can be applied to compress a given architecture. For example, quantization methods~\cite{yao2021hawq, liu2021post, courbariaux2016binarized} aim to reduce the required storage for data, often by substituting 32-bit floating-point numbers with 8-bit or 16-bit numbers or even utilizing binary values to represent the data. 
Pruning algorithms~\cite{lottery, guo2016dynamic,lee2018snip}, in their simplest form, remove parameters from a model to eliminate unnecessary redundancies within the network. Yet, more sophisticated algorithms may remove entire channels or filters from the network~\cite{FPGM,liu2019metapruning}. 
Knowledge distillation (KD) techniques~\cite{hinton2015distilling, gou2021knowledge} explore the concept of transferring knowledge from one model, referred to as the "teacher", to another, called the "student". The teacher represents a large pre-trained model with the desired knowledge, whereas the student denotes an untrained smaller model tasked with extracting knowledge from the teacher. However, as methods evolved, some algorithms~\cite{yuan2021revisiting, an2022efficient} modify the methodology by using the same network twice, eliminating the need for an extra teacher model. As these various compression methods progress, it is common to observe the adoption of two or more techniques, exemplified by the fusion of methods such as pruning and quantization in the same model.

Additionally, we discuss Neural Architecture Search (NAS) algorithms, a set of techniques designed to automate the model creation process while reducing human intervention. These algorithms autonomously search for optimal factors within a defined search space, such as network depth and filter settings. Research in this domain primarily focuses on refining the definition, traversal, and evaluation of the search space to achieve high accuracy without excessive time and resource consumption.

\vspace{-0.5em}
\subsection{Neural Network Deployment}
In Section~\ref{section:CNN Accelerator for Hardware system}, we navigate through the landscape of prevalent hardware accelerators dedicated to DL applications, including Graphics Processing Units (GPUs), Field-Programmable Gate Arrays (FPGAs), and Tensor Processing Units (TPUs). Moreover, we describe various dataflow types~\cite{chen2014diannao,tpu1,guo2017angel,lin2017data} and delve into data locality optimization methods~\cite{loopinter,looptiling,looptiling3}, exploring the intricate techniques that underpin efficient processing in DL workflows. The narrative further unfolds with a discussion of popular DL libraries~\cite{abadi2016tensorflow,paszke2019pytorch,chen2015mxnet} tailored for accelerating DL processes. This review encompasses the diverse tools and frameworks playing pivotal roles in optimizing the utilization of hardware accelerators. Additionally, we investigate co-designed solutions~\cite{wang2020sparse, parashar2017scnn, cho2021reconfigurable}, where achieving optimized and holistic results in accelerated DL requires careful consideration of hardware architecture and compression methods.

\vspace{-0.5em}
\subsection{Challenge and Future work}
Lastly, in Section~\ref{section:Challenge and Future work}, we embark on an exploration of emerging TinyML techniques designed to execute DL models on ultra-low-power devices, like MCUs, which typically consume less than 1 mW of power. Additionally, our paper delves into the intricacies of Large Language Models (LLMs), which present deployment challenges on devices with limited resources due to their enormous model sizes. As promising avenues in computer vision, deploying these methods on edge devices is crucial for widespread application.
\vspace{-0.5em}
\subsection{Contributions}
This paper aims to describe in a simple but accurate manner how lightweight architectures, compression methods, and hardware techniques can be leveraged to implement an accurate model in a resource-constrained device.
Our main contributions are summarized below:  
\begin{enumerate}
    \item Previous surveys only briefly reference a small number of works on lightweight architecture. We organize lightweight architectures into series, such as grouping MobileNetV1-V3 and MobileNeXt in the MobileNet series, and provide a history of lightweight architectures from their inception to the present.
    
    \item To cover the entire lightweight DL applications, we also cover the compression and hardware acceleration methods. Unlike many other surveys that do not explicitly establish connections between these techniques, our survey offers a thorough overview of each domain, providing a comprehensive understanding of their interconnections.

    \item As part of the forefront advancements in lightweight DL, we review the present challenges and explore future works. Firstly, we explore TinyML, an emerging approach engineered for deploying DL models on devices with remarkably constrained resources. Subsequently, we investigate various contemporary initiatives harnessing LLMs on edge devices, a promising direction in the realm of lightweight DL.
\end{enumerate}

\section{Lightweight Architecture Design}
\label{section:Lightweight Architecture Design}
To ease readers’ comprehension, we first introduce the fundamental knowledge of lightweight architecture, including the general metrics to estimate the computation cost of the NN and the widely used mechanisms of model compression.
Following that, we outline the lightweight CNN architecture and separate the sections by series, such as ShuffleNet and MobileNet series, according to their chronological order so that they can reflect the evolution of lightweight design and the advantage of its efficiency. Additionally, we discuss the efficient transformer, which offers a promising model capacity while maintaining a lightweight architecture.

\subsection{Prior Knowledge of Lightweight Architecture}
\subsubsection{Evaluation metrics for deep learning model}
In DL, the three most commonly used metrics for model compression are Floating Point Operations (FLOPs), Multiply-Accumulate Operations (MACs), and Memory Access Cost (MAC). FLOPs is the number of arithmetic operations the model performs on the floating points, including addition, subtraction, multiplication, and division~\cite{FLOPs_2021}. Similar to FLOPs, MACs also represent the total number of the floating point operations; however, MACs treat addition and multiplication as equivalent operations, in contrast to FLOPs, which distinguish between them~\cite{MAC_FLOPs}. Consequently, FLOPs $\approx$ 2$\times$MACs. On the other hand, MAC represents the amount of memory footprint of an NN, which corresponds to RAM usage~\cite{ma2018shufflenet}. 
Let $H$ and $W$ be the spatial size of the input and output feature maps for a convolution layer, $C_{in}$ is the number of input channels, $C_{out}$ is the number of output channels, and the kernel size is $k$,
\begin{align}
    MAC = H \cdot W (C_{in} + C_{out}) + k \cdot k (C_{in} \times C_{out}).
    \label{eq:MAC}
\end{align}
Specifically, the first and second terms of Eq.~\ref{eq:MAC} depict the memory footprint of the feature maps and weights for that particular convolution layer.

Furthermore, the most widely used metrics for measuring the inference speed of a model are throughput and latency. Throughput refers to the amount of data that can be processed or the number of tasks executed within a specified period. During inference, throughput is measured by the number of inferences per second. 
Latency is a measure of timing between the input data arriving at a system and the output data being generated and can be expressed in seconds per inference. 
The relationship between throughput and latency can be derived directly, and the detailed formula can be found in~\cite{evaluate}. 

\subsubsection{Pointwise convolution}
The pointwise convolution, also known as a $1 \times 1$ convolution, was first introduced in the inception module~\cite{szegedy2015going}. The inception module inserts the pointwise convolutions at the bottleneck to obtain deeper features with fewer FLOPs. 
Empowered by the adaptability of pointwise convolutions to accommodate modifications to the channel's dimensions, the Inception series of works was born~\cite{szegedy2016rethinking,szegedy2017inception,chollet2017xception}. Significantly, pointwise convolutions directly affect the model's computation time and the information richness of the architecture. 

\subsubsection{Group convolution}
\label{section:GroupConv}
The group convolution idea was proposed by AlexNet~\cite{krizhevsky2012imagenet}.
Group convolutions aim to divide the channels of feature maps into several groups and apply convolutions separately to each group. This process helps to reduce computational complexity by $N$ times, where $N$ represents the number of groups. 

However, there are still several shortcomings in group convolutions. Firstly, group assignments are fixed, and this factor restricts the information flow between groups, inevitably harming performance. Secondly, group convolutions cost additional MAC, especially when the number of groups is large, resulting in a much longer inference time. To solve the first problem, ShuffleNet~\cite{zhang2017shufflenet} shared group features to obtain deeper channel information. 
CondenseNet~\cite{huang2018condensenet} progressively prunes the unimportant connections using learned group convolutions (LGCs). 
Several works~\cite{wang2019fully, zhang2019differentiable} attempt to improve the original LGC to learn better optimal group structures. 
Furthermore, Dynamic Group Convolution (DGC)~\cite{su2020dynamic} highlights the importance of input channels via a salience generator and then uses a channel selector to assign groups adaptively. 

\subsubsection{Depthwise separable convolution}
The idea of a depthwise separable convolution was proposed in Xception~\cite{chollet2017xception}, which is the advanced version of the Inception family~\cite{szegedy2016rethinking,szegedy2017inception}. A depthwise separable convolution consists of a depthwise convolution followed by a pointwise convolution. According to the MAC, this is a computation-saving but time-consuming operation.
To address this issue, Tan et al.~\cite{Mixconv_2019} aggregate multiple kernel sizes into a single depthwise convolution and use AutoML~\cite{he2021automl} for navigating the search space. 

\vspace{-0.2em}

\subsection{Lightweight CNN Architecture}
\label{section:Lightweight Architecture of CNN}
\subsubsection{SqueezeNet series}
SqueezeNet series~\cite{iandola2016squeezenet,gholami2018squeezenext} is an early application to reduce parameters using pointwise convolution.
SqueezeNet~\cite{iandola2016squeezenet} proposes the fire module that constitutes the squeeze layer and the expand layer. The squeeze layer consists of pointwise convolution. It first squeezes features into lower dimensions and then passes them through an expansion layer, which separates the convolution operation into a pointwise convolution and a 3$\times$3 convolution. To solve the gradient vanishing problem and decrease the computation cost, SqueezeNext~\cite{gholami2018squeezenext} keeps the shortcut concept from ResNet~\cite{he2016deep} and decomposed 3$\times$3 kernel into two low-rank kernels, with sizes of 3$\times$1 and 1$\times$3. This augmented design reduces the parameters of the kernels from $k^{2}$ to $2k$, hence solving the inefficient problem of using depthwise separable convolutions.
Compared to AlexNet~\cite{krizhevsky2012imagenet}, SqueezeNet and SqueezeNext reduce the parameters by 50$\times$ and 112$\times$, respectively, while keeping AlexNet's level of accuracy on the ImageNet dataset. 


\subsubsection{ShuffleNet series}
\label{section:suffleSection}
The primary purpose of the ShuffleNet series~\cite{zhang2017shufflenet, ma2018shufflenet} is to improve the performance of group convolutions and the memory efficiency of depthwise separable convolutions.
After a group convolution, each group's output features form an individual channel, and performance suffers due to information not being shared between channels. To address this limitation, ShuffleNet~\cite{zhang2017shufflenet} applies a channel shuffle mechanism after the 1$\times$1 group convolution to facilitate cross-group information exchange so that features can maintain more global information channels.

ShuffleNetV2~\cite{ma2018shufflenet} investigates four practical guidelines to design a memory-efficient and lightweight model that can avoid heavy MAC problems. 
Firstly, equal input and output dimensions mean a smaller MAC. Secondly, MAC is large when groups are large, particularly for depthwise separable convolutions. Thirdly, it is best to avoid designing a wide network like the Inception series~\cite{szegedy2015going,szegedy2016rethinking,szegedy2017inception} because network fragments can result in a large MAC. Lastly, since element-wise manipulation in a network requires extra computation, avoiding it is efficient. 
This is often overlooked because it represents only a few FLOPs but increases MAC, as in depthwise separable convolutions.
 
\subsubsection{CondenseNet series}
\label{section:condenseSection}
Since shortcut connections effectively prevent the gradient vanishing problem~\cite{he2016deep}, some studies, such as DenseNet~\cite{huang2017densely} and the CondenseNet series~\cite{huang2018condensenet, yang2021condensenet} attempt to optimize NN structure based on shortcut connections.
DenseNet~\cite{huang2017densely} replaces the shortcut connections with dense connections, thus improving gradient flow within the bottleneck. 
Although dense connections increase the accuracy, CondenseNet~\cite{huang2018condensenet} observes that the magnitude of the connections far from the layers will decay exponentially, causing them to be heavy and slow. To this end, CondenseNet utilizes learned group convolutions (LGCs) to prune connections progressively. 
Before training, the filters are split into G groups of equal size.
Suppose $C_{in}$ is the number of input channels, $C_{out}$ is the number of output channels, and $F_{ij}^{g}$ denotes the kernel weights, including the weights of $j_{th}$ input, and $i_{th}$ output within a group $g\in G$. The importance of the $j_{th}$ incoming feature map for the filter group $g$ is computed by the averaged absolute value of weights between them across all outputs within the group, i.e., $\sum_{i=1}^{\frac{C_{out}}{G}} {|F_{ij}^{g}|}$, where columns in $F^g$ with small L1-norm value can be removed by zeroing them out. The structured sparsity within a group can be evaluated by applying the group-lasso regularizer~\cite{yuan2006model}, 

\begin{equation}   r=\sum_{g=1}^{G}\sum_{j=1}^{C_{in}}\sqrt{\sum_{i=1}^{\frac{C_{out}}{G}} {F_{ij}^{g}}^{2}}.
    \label{eq:CondenseNet}
\end{equation}
By using Eq.~\ref{eq:CondenseNet}, connections to less important features, represented by a small sparsity value, will be removed, resulting in effective pruning.

Recently, CondenseNetV2~\cite{yang2021condensenet} pointed out that the fixed connection mode limits the opportunities for feature reuse. To address this limitation, CondenseNetV2 aims to reactivate outdated features through a novel sparse feature reactivation module.  
In CondenseNetV2, the weight connections within each block are learned during training, as opposed to CondenseNet, which fixes the model's weight connections after pruning. As a result, this approach results in a performance gain by leveraging the underlying connection. 

\begin{figure*}[b]
\centering
\vspace{-0.5em}
\includegraphics[width=0.7\textwidth]{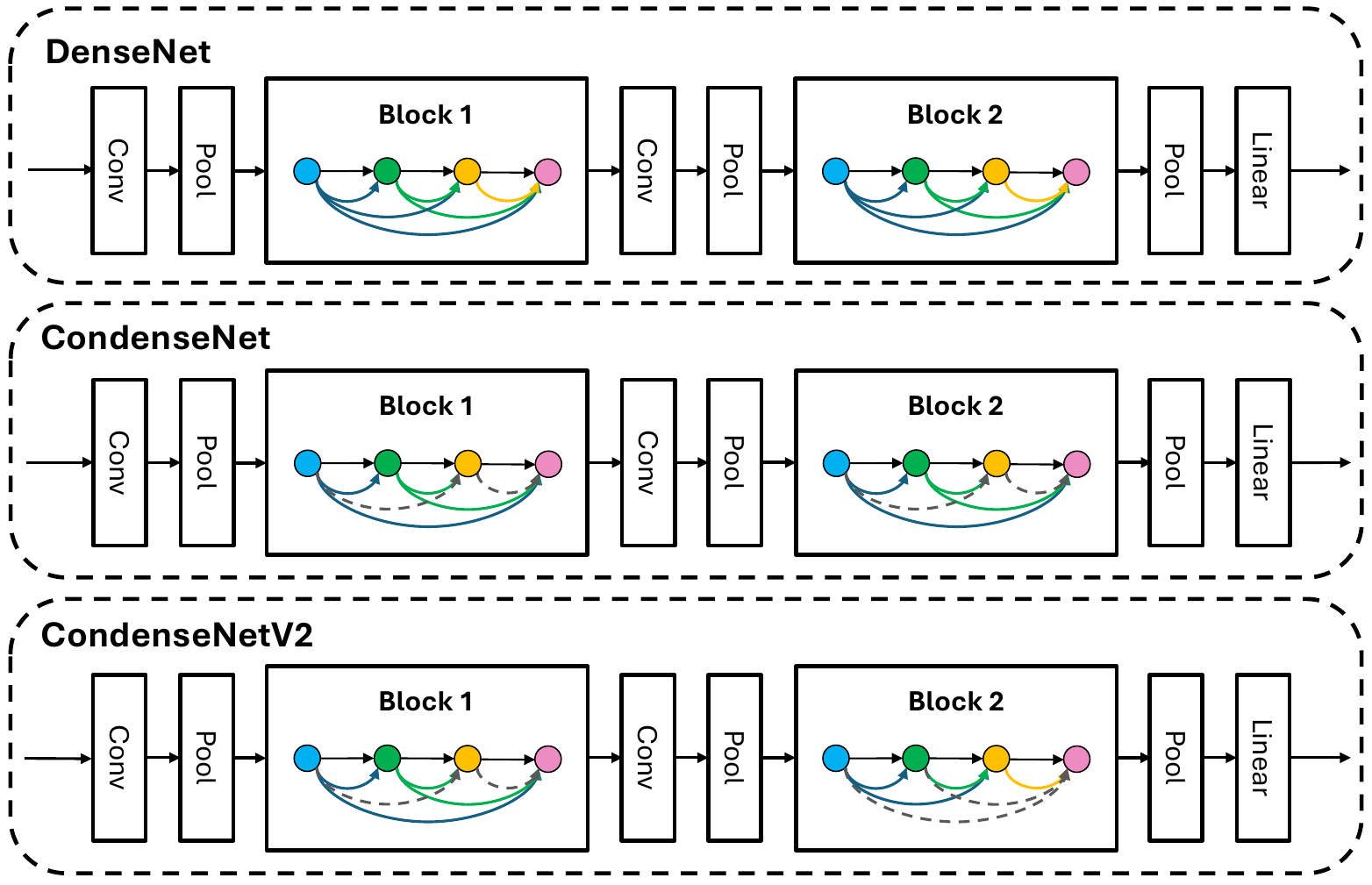}
\vspace{-0.5em}
\caption{Comparison of DenseNet, CondenseNet, and CondenseNetV2. 
Active weight connections are represented by solid color arrows, and pruned weight connections are represented by gray dashed arrows.}
\label{fig:Condenseconv}
\end{figure*}

Fig.~\ref{fig:Condenseconv} illustrates a graphical comparison, highlighting the architectural differences between DenseNet, CondenseNet, and CondenseNetV2. In DenseNet, weights between layers in a block are fully-connected, where weights in one layer are connected to weights in all other layers (solid-colored arrows). To make the network more efficient, CondenseNet uses LGCs to prune weight connections (gray dashed arrows), and once pruned, the connections for every block remain fixed, e.g., the connections in Block 1 and Block 2 are identical. CondenseNetV2 proposes a sparse feature reactivation mechanism to learn the connections' weights automatically during training. From Fig.~\ref{fig:Condenseconv}, we can observe that in Block 2 of CondenseNetV2, two pruned connections in Block 1 are reactivated while another two previously active connections in Block 1 are removed, demonstrating the dynamic nature of CondenseNetV2.

\subsubsection{MobileNet Series}
This series~\cite{howard2017mobilenets,sandler2018mobilenetv2,howard2019searching,zhou2020rethinking} includes prominent CNN models that can be deployed on IoT devices. Based on VGG~\cite{simonyan2014very} architecture, MobileNet~\cite{howard2017mobilenets} applies depthwise separable convolutions to create an efficient model, which is shown to perform significantly faster across a broad range of tasks and applications.
Discovering that ReLU activations can lead to severe information loss of features with lower dimensions, MobileNetV2~\cite{sandler2018mobilenetv2} replaces the ReLU activation with a linear combination in the last layer of the residual bottleneck to mitigate the information loss. In addition, MobileNetV2 introduces an inverted residual block, where the number of channels is first increased and then recovered in the residual bottleneck, improving the accuracy. Shortcut connections~\cite{he2016deep} are also added to enhance the gradient propagation.

NetAdapt~\cite{yang2018netadapt} applies layer-wise optimization to simplify the network and to achieve high accuracy within limited hardware resources.
Based on this, MobileNetV3~\cite{howard2019searching} leverages platform-aware NAS~\cite{tan2019mnasnet} to optimize the block-wise structure and implements SENet~\cite{hu2018squeeze} (channel attention module) in the bottleneck structures, resulting in better accuracy. To reduce MAC and establish a quantization-friendly network, ReLU is replaced with H-swish activation. As an alternative to the inverted residual block, MobileNeXt~\cite{zhou2020rethinking} develops a Sandglass block by flipping the inverted residual block to enhance the transmission of wider architectures since wider layers might lead to more gradient confusion, making model training harder.

\subsubsection{Shift-based series}   
CNN is computationally expensive due to many multiplication and addition operations. ShiftNet~\cite{wu2018shift} pioneered the replacement of spatial convolutions with Group Shift convolution. Unlike standard convolutions, shift convolutions only perform shifting operations on feature maps and apply padding to those offset areas. In contrast to multiplication operations, shift convolution can achieve zero parameters and FLOPs, thus drastically reducing their number.

\begin{figure}%
\centering
\subfigure[Group Shift]{
\label{fig:GroupedShift}%
\includegraphics[width=0.15\textwidth]{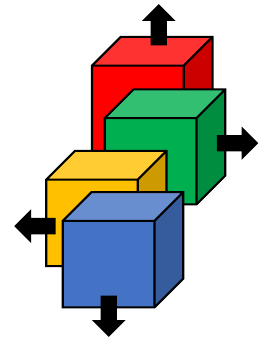}}
\qquad
\subfigure[Active Shift]{
\label{fig:ActiveShift}
\includegraphics[width=0.18\textwidth]{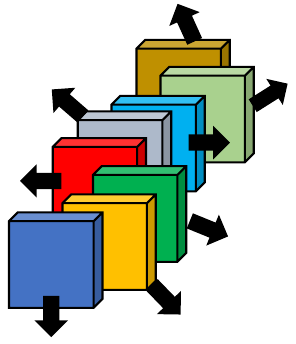}}
\subfigure[Sparse Shift]{
\label{fig:SparseShift}
\includegraphics[width=0.2\textwidth]{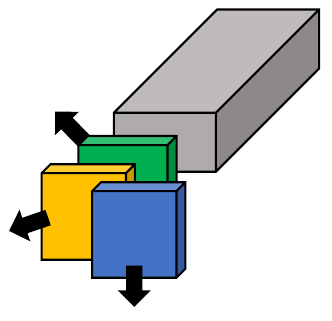}}
\vspace{-1em}
\caption{The variant of Shift-based convolution~\cite{chen2019all}.}
\label{fig:shifts}
\vspace{-0.5em}
\end{figure}

Some studies attempt to improve the performance based on shift convolution layers. For example, Jeon et al.~\cite{jeon2018constructing} propose an Active Shift Layer that makes shifts learnable instead of heuristic assignments.
Chen et al.~\cite{chen2019all} point out that because the number of shifts is fixed, implementing them requires a lot of trial and error, limiting the network's functionality. Thus, they propose a Sparse Shift Layer to eliminate meaningless memory movement. The non-shift channels remain the same. Fig.~\ref{fig:shifts} compares these three shift operations.

AddressNet~\cite{he2019addressnet} observes that a smaller amount of parameters or computation (FLOPs) does not always lead to a direct reduction in inference time, even with shift convolution's zero parameters and zero FLOPs~\cite{wu2018shift}. To optimize the speed of GPU-based machines, AddressNet changes channel shuffle~\cite{zhang2017shufflenet} to channel shift since channel shuffle produces additional memory space and time-consuming permutations, further eliminating the redundant direction. Similar to AdderNet~\cite{Chen_2020_CVPR}, DeepShift~\cite{elhoushi2021deepshift} is constructed solely with addition operations, replacing all multiplications with bit-wise shifts and sign flips, significantly reduces the operation time and energy consumption.

\subsubsection{Add-based Series}  
Multiplication and addition operations constitute many convolution operations, resulting in extra calculations. AdderNet~\cite{Chen_2020_CVPR} attempts to exclusively use additions, using an L1-norm distance as a response criterion between filters and feature maps. This operation is known as Absolute-difference-accumulation~\cite{add_hard}, and it accelerates the network and allows the reuse of computation results in order to reduce energy consumption.

You et al.~\cite{you2020shiftaddnet} introduce ShiftAddnet, focusing more on hardware efficiency. ShiftAddnet proposes a new metric for performance comparison, expressive capacity, which refers to the accuracy achieved by the model under similar hardware conditions. Experimental results show that shift-based networks~\cite{wu2018shift,jeon2018constructing,chen2019all,he2019addressnet,elhoushi2021deepshift} provide greater hardware efficiency but have a lower expressive capacity than multiplication-based networks. Conversely, the fully additive network~\cite{Chen_2020_CVPR} is inefficient since repeated additions are used to replace multiplications, although it can achieve better accuracy. Therefore, ShiftAddnet combines the benefits of bit-wise shifts~\cite{elhoushi2021deepshift} and the efficiency of additive networks~\cite{Chen_2020_CVPR} to achieve state-of-the-art results on two IoT datasets: FlatCam~\cite{FlatCam} and Head Pose~\cite {Headpose}.

\subsubsection{EfficientNet Series}   
Almost all networks attempt to improve performance by adjusting depth, width, and resolution. To achieve the best performance and lightweight combination, it is crucial to pick the right combination. EfficientNet~\cite{tan2019efficientnet} proposes a simple grid search algorithm, compound scaling, to seek scaling factors (depth, width, and resolution), achieving accuracy with lower computation costs. EffectiveNetV2~\cite{tan2021efficientnetv2} proposes a training-aware NAS to find a good trade-off for accuracy $A$, training speed $S$, and parameters $P$. It uses a search reward formulated as a simple weighted product, $A \cdot S^{w} \cdot P^{v}$, where $w = -0.07$ and $v = -0.05$ are empirically determined to balance the trade-off. To address the inefficiency of depthwise convolution, EfficientNetV2 replaces stage 1-3 MBConv~\cite{sandler2018mobilenetv2} with Fused-MBConv~\cite{gupta2020accelerator} in its architecture design, offering better performance and trade-off in terms of accuracy, parameters, and FLOPs. Besides, for a more robust network, EfficientNetV2 selects adaptive regulation during the training process because using identical regularization terms for images of different resolutions is inefficient.

\subsubsection{Discussion and Summary}
Table~\ref {table:summaryLightweightArch} compares the performance of lightweight CNN architectures on the ImageNet dataset. The horizontal lines separate the models of different series. 
From the table, we can observe that there is no one-size-fits-all architecture. Oftentimes, it is a trade-off between accuracy and efficiency. For example, AddressNet-20 maximizes efficiency at the expense of accuracy. Conversely, the most accurate variants of the EfficientNet series are among the least efficient ones. Drawing from this analysis, we provide recommendations on selecting the suitable models and hardware.

\noindent \textbf{How to choose an adequate lightweight model and compatible hardware?} 
The first crucial step is to check lightweight models' specifications and hardware compatibility. For example, depthwise separable convolutions have huge MAC and high RAM requirements. It is, therefore, imperative to employ a network on hardware that considers both RAM and storage capacity. To this end, Fan et al.~\cite{co-op1} redesign the depthwise separable convolution and channel shuffle modules to be hardware-friendly on FPGA.
Moreover, to minimize the inference time and to support deployment on a small target device, replacing multiplication with shift or add operations can effectively reduce the total parameters and MACs/FLOPs. 
Thus, ShiftNet or AdderNet series can be a good choice since they require smaller parameters and MACs. Within these two series, AddressNet-20 gives the best performance. For target devices with relatively more storage, such as mobile phones or GPUs, models with higher accuracy are preferred for a better user experience. EfficientNetV2-L can thus be considered since it achieves the highest Top-1 accuracy. However, it is important to note that the EfficientNet series costs a disproportionately higher number of parameters and MACs, which limits the application under low-end devices. Another way to achieve a better trade-off model is to apply fundamental compression methods such as pruning, quantization, and NAS~\cite{chen2019detnas,tan2019mnasnet} (see Section 3) to adjust the architecture. This can be an efficient technique to reduce MACs/FLOPs, parameters, and inference time.

Some lightweight methods, such as SqueezeNet and ShuffeNet, may not be able to take full advantage of GPU-accelerated performance due to the lack of customized designs~\cite{add_hard}.
Additionally, if pruning is applied to a network, like the CondenseNet series, the network structure might be irregular, preventing the target device from supporting it. In such a scenario, parallelism requires specifically designed computing hardware.
Fortunately, customized hardware can be designed to fit a lightweight model. For instance, Um et al.~\cite{add_hard} note that CIM is incompatible with AdderNet because it cannot predict details of an absolute difference nor reuse the computation results. Therefore, they designed a novel ADA-CIM processor offering low-cost sign prediction and higher energy efficiency.

\begin{table}
\footnotesize
\caption{Comparison of Lightweight CNN Architectures on the ImageNet dataset. Note that we use~\textbf{bold} to emphasize the models with the best accuracy, least parameters, and lowest MACs, with the respective values being also~\underline{underlined} for enhanced readability.}
\vspace{-0.5em}
\label{table:summaryLightweightArch}
\begin{tabular}{|l|c|c|c|c|} 
\hline
\multicolumn{1}{|c|}{\textbf{Model}} & \textbf{Top-1} & \textbf{Top-5} & \textbf{Params. (M)} & \textbf{MACs (G)}  \\ 
\hline
AlexNet~\cite{krizhevsky2012imagenet}                               & 57.1           & 80.3           & 60.9               & 0.725              \\
ResNet-50~\cite{he2016deep}                             & 76.0           & 93.0           & 26.0               & 4.100              \\ 
\hline
SqueezeNet~\cite{iandola2016squeezenet}                            & 57.5           & 80.3           & 1.2                & 0.837              \\
SqueezeNext~\cite{gholami2018squeezenext}                           & 59.1           & 82.6           & 0.7                & 0.282              \\ 
\hline
ShuffleNetV1-1.5~\cite{zhang2017shufflenet}                      & 71.5           & -              & 3.4                & 0.292              \\
ShuffleNetV2-1.5~\cite{ma2018shufflenet}                      & 72.6           & 90.6           & 3.5                & 0.299              \\ 
\hline
1.0-MobileNetV1~\cite{howard2017mobilenets}                       & 70.6           & -              & 4.2                & 0.569              \\
MobileNetV2-1.4~\cite{sandler2018mobilenetv2}                       & 74.7           & -              & 6.9                & 0.585              \\
MobileV3-S~\cite{howard2019searching}                            & 67.4           & -              & 2.5                & 0.056              \\
MobileV3-L~\cite{howard2019searching}                            & 75.2           & -              & 5.4                & 0.219              \\
MobileNeXt-1.0~\cite{zhou2020rethinking}                        & 74.0           & -              & 3.4                & 0.300              \\
\hline
ShiftResNet-20~\cite{wu2018shift}                        & 68.6           & -              & 0.2                & 0.046              \\
ShiftResNet-56~\cite{wu2018shift}                        & 72.1           & -              & 0.6                & 0.102              \\
ShiftNet-A~\cite{wu2018shift}                            & 70.1           & 89.7           & 4.1                & 1.400              \\
ShiftNet-B~\cite{wu2018shift}                            & 61.2           & 83.6           & 1.1                & 0.371              \\
FE-Net-1.0~\cite{chen2019all}                            & 72.9           & -              & 3.7                & 0.301              \\
FE-Net-1.37~\cite{chen2019all}                          & 75.0           & -              & 5.9                & 0.563              \\
\textbf{AddressNet-20}~\cite{he2019addressnet}  & 68.7 & -  & \textbf{\underline{0.1}}  & \textbf{\underline{0.022}} \\
AddressNet-44~\cite{he2019addressnet}                         & 73.3           & -              & 0.2                & 0.053              \\ 
\hline
AdderNet-Resnet18~\cite{Chen_2020_CVPR}                     & 67.0           & 87.6           & 3.6                & -                  \\
AdderNet-Resnet50~\cite{Chen_2020_CVPR}                     & 74.9           & 91.7           & 7.7                & -                  \\ 
\hline
DenseNet-169~\cite{huang2017densely}                          & 76.2           & 93.2           & 14.0               & 3.500              \\
DenseNet-264~\cite{huang2017densely}                          & 77.9           & 93.9           & 34.0               & 6.000              \\
CondenseNet~\cite{huang2018condensenet}                           & 71.0           & 90.0           & 2.9                & 0.274              \\
CondenseV2-A~\cite{yang2021condensenet}                          & 64.4           & 84.5           & 2.0                & 0.046             \\
CondenseV2-B~\cite{yang2021condensenet}                          & 71.9           & 90.3           & 3.6                & 0.146              \\ 
\hline
EfficientNet-B1~\cite{tan2019efficientnet}                       & 79.2           & 94.5           & 7.8                & 0.700              \\
\textbf{EfficientNet-B7}~\cite{tan2019efficientnet}                       & 84.4           & \textbf{\underline{97.1}}          & 66.0               & 37.000             \\
EfficientNet-X-B7~\cite{li2021searching}                        & 84.7           & -              & 73.0               & 91.000             \\
EfficientNetV2-S~\cite{tan2021efficientnetv2}                      & 83.9           & -              & 24.0               & 8.800              \\
EfficientNetV2-M~\cite{tan2021efficientnetv2}                      & 85.1           & -              & 55.0               & 24.000             \\
{\textbf{EfficientNetV2-L~\cite{tan2021efficientnetv2}}}  & \textbf{\underline{85.7}}& -   & 121.0 & 53.000  \\
\hline
\end{tabular}
\end{table}
\vspace{-0.5em}

\subsection{Transformer-based Series}   
\begin{figure}[b] 
\centering 
\includegraphics[width=0.75\columnwidth]{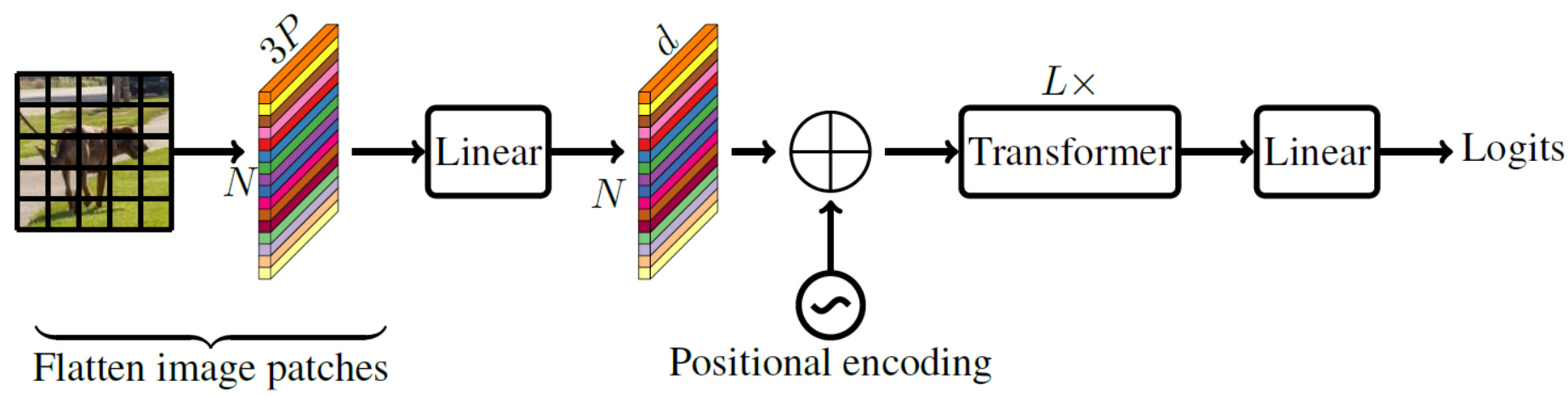} 
\vspace{-0.5em}
\caption{Standard Vision Transformer, where $P= h \times w$, $h,w$ represents the height and the width of the images. $N$ is the number of image patches, $L$ is the number of transformer blocks, and $d$ is the dimension~\cite{mehta2021mobilevit}.} 
\label{fig:transformer}
\end{figure}

Transformer models are widely used in NLP~\cite{vaswani2017attention} and have recently obtained promising results in computer vision tasks~\cite{liu2021swin,liu2021swinv2,zhang2022dino}. 
Fig.~\ref{fig:transformer} shows the architecture of a typical vision transformer.
Transformers are notable for having a significant drawback in that they require a large number of parameters and a high MAC to maintain their performance, which results in a significant amount of time needed for both the training and inference phases, particularly when the input sequence is long. Additionally, the computation and network structures inside transformers are more complex than those of CNNs. The huge number of FLOPs and parameters make practical inference and hardware deployment more difficult. To bridge the gap between transformers and real-world applications, efficient transformers will be discussed in the following sub-sections.

\vspace{-0.8em}

\subsubsection{Lite attention module}
To address the heavy MAC and huge computation requirements in the self-attention layers, Long-Short Range Attention (LSRA)~\cite{wu2020lite} was proposed to extract the global and local features separately, alleviating the attention computations in the feed-forward network (FFN). Child et al.~\cite{child2019generating} effectively exploit stride and fixed operations to form a sparse connectivity matrix. Linformer~\cite{wang2020linformer} decomposes self-attention into several low-rank matrices using linear projection, reducing the complexity of self-attention from $N^{2}$ to $N$, where $N$ denotes the sequence length. Choromanski et al.~\cite{choromanski2020rethinking} proposes a linear self-attention mechanism based on the FAVOR+ (fast attention with positive orthogonal random features) approach to construct an approximate softmax operation. FAVOR+ enables unbiased estimation of self-attention with low estimation variance, reducing spatial and temporal complexity. Reformer~\cite{kitaev2020reformer} utilizes locality-sensitive hashing to replace dot product operations in attention. It directly decreases the computation requirements from $N^{2}$ to $Nlog(N)$, allowing longer sequence inputs to be considered. In addition, Reformer employs a reverse residual layer~\cite{gomez2017reversible} to save GPU memory by $L$ times (number of layers). Unlike traditional residuals, a reverse residual layer does not require activation data to be stored in each layer. The complexity of these efficient transformers is depicted in Table~\ref{table:efficientTransformer}. 

In addition, transformers stack many FFNs to obtain better-integrated features. Specifically, an FFN is a series of linear transformations that require a lot of calculations due to its dense connections. To tackle this issue, Mehta et al.~\cite{mehta2018pyramidal} introduce grouped linear transformations (GLTs), which incorporate the concept of group convolution to make the transformer block more lightweight. Facing the same shortcoming from the group convolutions (as presented in Section~\ref{section:GroupConv}), the hierarchical group transformation (HGT)~\cite{mehta2019define} aims to enhance the information flow between groups using a split layer and a skip connection operation. DeLighT~\cite{mehta2020delight} exploits GLTs to make feature dimensions wider and deeper, making it possible to use single-head attention instead of multi-head attention. This technique decreases the computation cost in the attention operation from $d_{m}N^{2}$ to $d_{o}N^{2}$, where $d_{m}$, and $d_{o}$ are the input dimension and output dimension respectively. 

\begin{center}
\bgroup
\def\arraystretch{1.2}%
\begin{table}[b!]
\footnotesize
\caption{The complexity of Efficient Transformers~\cite{wang2020linformer}.}
\vspace{-1em}
\label{table:efficientTransformer}
\begin{tabular}{|l|c|c|llllllllll} 
\cline{1-3}
\multicolumn{1}{|c|}{\textbf{Model}} & \textbf{Complexity per Layer} & \textbf{Sequential Operation}  \\ 
\hline
Transformer~\cite{vaswani2017attention}                          & $O(N\textsuperscript{2})$       & $O(N)$                           \\
Sparse Transformer~\cite{child2019generating}                    & $O(N\sqrt{N})$                & $O(1)$                           \\
Linformer~\cite{wang2020linformer}                            & $O(N)$                         & $O(1)$                           \\
Reformer~\cite{kitaev2020reformer}                             & $O(Nlog(N))$                   & $O(log(N))$                      \\
\hline
\end{tabular}
\end{table}
\egroup
\end{center}

\vspace{-2.0em}
\subsubsection{Token sparsing} 
Vision transformer (ViT)~\cite{dosovitskiy2020vit} is the earliest work that applied transformers to solve an image classification task. It first splits an image into several patches and flattens them so that it can be passed in as an embedding sequence input to the transformer architecture. As the resolution of images in ImageNet is 224x224, their tokens require significantly more computation compared to other datasets with smaller resolutions, such as CIFAR-10 and CIFAR-100 (32x32).

To address this, T2T-ViT~\cite{yuan2021tokens} observes that image splitting in transformers causes a loss of local relationships between tokens since there is no overlap between the tokens. Hence, they employ soft unfolding to combine the surrounding spatial tokens into high-dimensional manifolds, enabling smaller MLP sizes and increasing memory efficiency. 

An extensive study on transformers~\cite{naseer2021intriguing} demonstrates that transformers are robust to patch drops, with only a slight decrease in accuracy when patches suffer from distortion or occlusions. DynamicViT~\cite{rao2021dynamicvit} integrates a prediction module between transformer blocks to mask the less significant tokens. The prediction module is a binary decision mask in the range (0,1) that measures the importance of tokens. EViT~\cite{liang2021evit} computes attentiveness scores from class tokens and other tokens and keeps top-K tokens, representing the highest positive correlation to the prediction. A-ViT~\cite{yin2021adavit} adaptively changes the number of tokens at different depths based on the complexity of the input image to reduce the inference time in ViT.

\subsubsection{Lightweight hybrid models} 
Due to the long-range dependence property inherent in attention mechanisms, transformer networks outperform CNN in accuracy. However, a transformer network lacks strong inductive biases~\cite{dai2021coatnet,graham2021levit,liu2021swin}, making it difficult to train and requires extra data augmentation and heavy regularization to maintain performance~\cite{touvron2021training}. On the other hand, CNN extracts features based on sliding windows, resulting in stronger inductive biases, which make models easier to train and have better generalizability. Interestingly, the aggregation of CNN and transformer networks~\cite{wu2021cvt,srinivas2021bottleneck,d2021convit,xiao2021early} produces versatile and powerful models. Since the hybrid models would have many parameters, DeiT~\cite{touvron2021training} applies KD and achieves better accuracy with less latency than CNN under comparable parameters. To improve data efficiency and simplify model complexity, the student model, a ViT model, added a distillation token to provide insight into the inductive biases of a CNN-based teacher model. MobileViT~\cite{mehta2021mobilevit} points out that transformer-based networks perform worse than CNN networks under similar parameters because they are still bulky. MobileViT employs MobileNetV2~\cite{sandler2018mobilenetv2} as the CNN backbone to obtain inductive biases and replaces the MBconv block in MobileNetV2 with a MobileViT block with unfolding and folding operations, which can compute long-range dependencies like a transformer. Similarly, MobileFormer~\cite{chen2021mobile} devises a parallel structure consisting of CNNs and transformers to achieve feature fusion. Inductive bias and the ability to capture global features are incorporated via two-way cross-attention.

\subsubsection{Discussion and Summary}
Recent transformer models focus on lighter and more powerful architectures. This observation is apparent from Table~\ref{table:summaryTransformers}, where many recent transformers, such as T2T-ViT~\cite{yuan2021tokens} and DymViT-LVit~\cite{rao2021dynamicvit}, are shown to achieve higher accuracy with significantly fewer parameters and lower FLOPS than the original ViT and ResNet-based CNNs. Specifically, we split the discussion into 3 sub-sections with bold headings.\\
\noindent \textbf{VIT \& KD transformer.}
Inspired by~\cite{hinton2015distilling}, several papers~\cite{touvron2021training, kd_targetaware, reviewkd} apply KD to distill the inductive bias from the CNN-based teacher models to the transformer-based student models. For example, the design of DeiT-B~\cite{touvron2021training} architecture integrates a CNN-based teacher model, a RegNetY-16G~\cite{regnet}, and a transformer-based student model, ViT-B~\cite{dosovitskiy2020vit}. Results show that DeiT-B outperforms all the models in terms of Top-1 accuracy, achieving an accuracy of 84.5\%. Despite their stronger abilities, transformer-based student models require a large network to maintain their performance since they are harder to converge than CNN models~\cite{dai2021coatnet}. \\
\noindent \textbf{VIT \& CNN hybrid transformer.}
To overcome the shortcomings of the KD-based transformer models, the hybrid models~\cite{dai2021coatnet, mehta2021mobilevit, chen2021mobile} utilize both the convolution and transformer layers in the network. By doing so, they can obtain stronger inductive bias, leading to better convergence during training. Thus, hybrid models typically have fewer FLOPs and parameters. For example, Mobile-Former-96M~\cite{chen2021mobile} achieves the lowest FLOPs of 0.096G while MobileViT-XS~\cite{mehta2021mobilevit} has the lowest parameters, which is 2.3 M. These hybrid models are extremely lightweight but sometimes, efficiency is achieved at the expense of accuracy, as we can observe from their performance in Table \ref{table:efficientTransformer}. For instance, MobileVit-XS has roughly half the total parameters of MobileViT-S, its counterpart, but its accuracy has significantly dropped by 3.6\%. Another noteworthy observation shows that although Mobile-Former-96M achieves the lowest FLOPS, its parameter size was doubled, and accuracy is 2.0\% lower compared to MobileVit-XS. This demonstrates that there is not always a correlation between FLOPs and total parameters and that lowering FLOPs appears to have a greater impact on accuracy than lowering parameters.

\noindent \textbf{VIT \& Token sparsing transformer.}
Another series of efficient transformers~\cite{yuan2021tokens,rao2021dynamicvit,naseer2021intriguing,liang2021evit,yin2021adavit} aim to prune the transformer structure efficiently via token sparsing. From the results, token sparsing-based models achieve a competitive accuracy with fewer parameters and FLOPs. It is worth noting that EViT-DeiT-S (k=0.7)~\cite{liang2021evit} can reach the highest throughput, 5408 images per second. Therefore, for a faster transformer-based model, such as accomplishing a real-time system, aggregating tokens into smaller amounts may provide a promising solution. 

Due to their competitive accuracy and lightweight design~\cite{kang2023,luo2022towards}, lightweight transformer models are gaining popularity in a wide range of applications, such as edge AI and mobile AI; more details of efficient transformers can be found in~\cite{han2022survey,tay2020efficient}.

\vspace{-0.2em}

\begin{table}
\centering
\footnotesize
\caption{Comparison of Lightweight Transformer Models on the ImageNet dataset. We use~\textbf{bold} to emphasize the models with the least parameters, highest throughput, lowest FLOPs, and best accuracy, with the corresponding values also~\underline{underlined} for enhanced readability.}
\vspace{-1em}
\label{table:summaryTransformers}
\begin{tabular}{|c|l|c|c|c|c|c|} 
\cline{1-7}
\textbf{Categories}     & \multicolumn{1}{c|}{\textbf{Model}} & \begin{tabular}[c]{@{}c@{}}\textbf{Image}\\\textbf{(Size)}\end{tabular} & \textbf{Params. (M)} & \begin{tabular}[c]{@{}c@{}}\textbf{Throughput}\\\textbf{(image/s)}\end{tabular} & \textbf{FLOPs(G)} & \begin{tabular}[c]{@{}c@{}}\textbf{ImageNet}\\\textbf{Top-1}\end{tabular}    \\ 
\cline{1-7}
\multirow{4}{*}{CNN}  & ResNet50~\cite{he2016deep}    & 224$\times$224   & 25.5    & -    & 4.13   & 76.2   \\
& ResNet101~\cite{he2016deep}    & 224$\times$224  & 44.6     & -    & 7.9   & 77.4    \\
& ResNet152~\cite{he2016deep}   & 224$\times$224  & 60.2      & -    & 11.0   & 78.3    \\ 
& RegNetY-16GF~\cite{regnet}   & 224$\times$224  & 84.0     & 334.7    & -  & 82.9   \\ 
\cline{1-7}
\multirow{2}{*}{ViT}    & ViT-B/16~\cite{dosovitskiy2020vit}      & 384$\times$384     & 86.6    & 85.9    & 17.6   & 77.9 \\\
& ViT-L/16~\cite{dosovitskiy2020vit}    & 384$\times$384   & 307.0   & 27.3    & 63.6   & 76.5  \\ 
\cline{1-7}
\multirow{5}{*}{ViT \& KD}    & DeiT-Ti~\cite{touvron2021training}  & 224$\times$224    & 5.0     & 2536.5     & -    & 72.2 \\
& DeiT-Ti~\cite{touvron2021training}  & 224$\times$224  & 6.0  & 2529.5   & -  & 74.5  \\
& DeiT-S~\cite{touvron2021training}   & 224$\times$224  & 22.0   & 936.2  & 4.6  & 81.2  \\
& DeiT-B~\cite{touvron2021training}   & 224$\times$224  & 87.0   & 290.9 & 17.6   & 83.4 \\
& \textbf{DeiT-B}~\cite{touvron2021training}   & 384$\times$384  & 87.0   & 85.8   & 17.6  & \underline{\textbf{84.5}} \\ 
\cline{1-7}
\multirow{11}{*}{\begin{tabular}[c]{@{}c@{}}ViT \& \\ Token Sparsing
\end{tabular}} 
& T2T-ViT-14~\cite{yuan2021tokens}   & 224$\times$224  & 21.5  & -  & 5.2   & 81.5  \\
& T2T-ViT-14~\cite{yuan2021tokens}   & 384$\times$384  & 21.5  & -   & 17.1  & 83.3  \\
& T2T-ViT-19~\cite{yuan2021tokens}   & 224$\times$224  & 39.2  & -  & 8.9    & 81.9  \\
& DymViT-LViT-S/0.5~\cite{rao2021dynamicvit}  & 224$\times$224    & 26.9    & -   & 3.7   & 82.0  \\
& DymViT-LViT-M/0.7~\cite{rao2021dynamicvit}  & 224$\times$224    & 57.1    & -   & 8.5   & 83.8  \\
& EViT-DeiT-S (k=0.5)~\cite{liang2021evit}    & 224$\times$224    & 22.0    & 4385   & 3.0    & 79.5  \\
& \textbf{EViT-DeiT-S (k=0.7)}~\cite{liang2021evit} & 224$\times$224  & 22.0  & \underline{\textbf{5408}} & 2.3 & 78.5 \\
& EViT-LViT-S (k=0.5)~\cite{liang2021evit}    & 224$\times$224    & 26.2    & 3603 & 3.9    & 82.5  \\
& EViT-LViT-S (k=0.7)~\cite{liang2021evit}    & 224$\times$224  & 26.2   & 2954   & 4.7   & 83.0 \\
& A-ViT-T~\cite{yin2021adavit}     & 224$\times$224    & 5.0   & 3400   & 0.8   & 71.0  \\
& A-ViT-S~\cite{yin2021adavit}     & 224$\times$224    & 22.0  & 1100  & 3.6  & 78.6   \\ 
\cline{1-7}
\multirow{5}{*}{\begin{tabular}[c]{@{}c@{}}ViT \& CNN\\(Hybrid models)\end{tabular}} 
& \textbf{Mobile-Former-96M~\cite{chen2021mobile}}  & 224$\times$224   & 4.6   & -   & \underline{\textbf{0.096}}  & 72.8  \\
& Mobile-Former-29\cite{chen2021mobile}  & 224$\times$224     & 11.4    & -  & 0.294  & 77.9 \\
& Mobile-Former-508M~\cite{chen2021mobile}  & 224$\times$224     & 14.0    & -  & 0.508  & 79.3 \\
&\textbf{MobileViT-XS~\cite{mehta2021mobilevit}}  & 224$\times$224  &  \underline{\textbf{2.3}}  & -  & 0.7  & 74.8  \\
& MobileViT-S~\cite{mehta2021mobilevit}     & 224$\times$224     & 5.6     & -   & -    & 78.4  \\
\cline{1-7}
\end{tabular}
\end{table}

\section{Fundamental methods in model compression}
\label{section:Fundamental methods in model compression}
In this section, we explore popular compression methods used in recent years and their improvements over time. These techniques encompass pruning~\cite{obd, obs, lottery, FPGM, cp2023}, quantization~\cite{dong2019hawq, Adaptive_Quant_2020, courbariaux2016binarized}, knowledge distillation~\cite{hinton2015distilling, gou2021knowledge, zhang2017deep}, and neural architecture search~\cite{liu2018darts, wu2019fbnet}, which are widely adopted for designing efficient models. We further unveil a detailed exploration of each method, offering deeper insights that stem from their distinctive characteristics.

\vspace{-0.5em}
\subsection{Pruning}
DL models frequently comprise numerous learnable parameters, requiring extensive training. Pruning methods aim to compress and expedite NNs by removing redundant weights. These pruning methods can be categorized as either unstructured or structured.

\subsubsection{Unstructured pruning}
Unstructured pruning aims to identify and eliminate individual weights from the network, regardless of where they are located. This method imposes no restrictions or rules on weight trimming. Specifically, the nodes with the removed weights are not physically removed from the network; instead, the weights are set to zero. Since this operation results in numerous zero multiplications, models can be significantly compressed for faster inference. As illustrated in Fig.~\ref{fig:structuredpruning} (left), unstructured pruning may cause the pruned network to have an irregular structure. Early works in pruning, such as Optimal Brain Damage~\cite{obd} and Optimal Brain Surgeon~\cite{obs}, utilize second-order derivatives and Hessian matrices to assess the importance of weights in the network and subsequently prune them. While these methods demonstrate impressive performance, they demand substantial computational power.

To this end, Dong et al.~\cite{OPS-1} introduce a method that restricts the computation of second-order derivatives. This approach does not require the computation of the Hessian matrix over all parameters; instead, it focuses on specific layers of the model. Similarly, Frankle et al.~\cite{lottery} propose the lottery ticket hypothesis, where they attempt to find more manageable and pruned sub-networks while maintaining a performance comparable to the original network. In their approach, they prune the nodes, subsequently restoring the original pre-training initialization values of the untouched nodes, and repeat this cycle until a certain level of sparsity is achieved.

However, unstructured pruning can significantly reduce accuracy when weights are pruned during the training process before the network converges. Unfortunately, the pruned connections cannot be restored. To address this limitation, Guo et al.~\cite{guo2016dynamic} introduce a splicing algorithm capable of recovering previously deleted connections that are discovered to be important at any point in time. Furthermore, Namhoon et al.~\cite{lee2018snip} propose a single-shot network pruning approach in which they prune the network before the training begins. Instead of analyzing the model's final weights after training, they examine the response of the loss function to variance scaling during initialization. This innovative approach allows the network to be pruned just once before training, providing a more convenient and effective pruning method.

\subsubsection{Structured Pruning}
\begin{figure}[b] 
\centering
        \includegraphics[width=\textwidth]{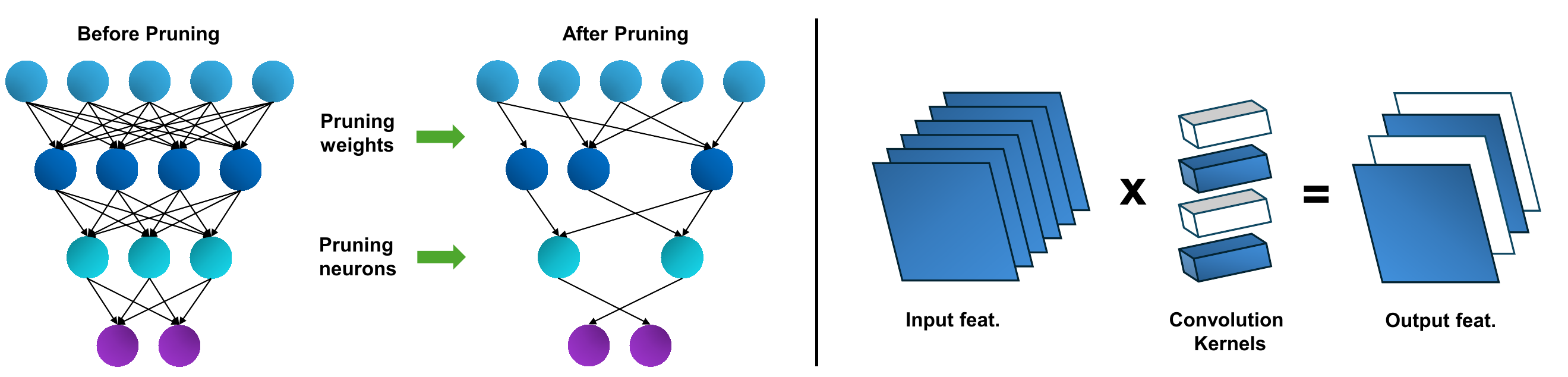}
	\caption{Illustration of pruning methods: unstructured pruning (left), and structured pruning (right). Pruned components are shown in white color. Take note of the change in the pruned component's output dimensions.} 
	\label{fig:structuredpruning}
    \vspace{-1em}
\end{figure}

Structured pruning methods remove pruned components from a pre-trained network and preserve its regular structure, as shown in Fig.~\ref{fig:structuredpruning} (right). Common structured pruning methods include filter pruning~\cite{FPGM, SFP, LFPC, ASTER} and channel pruning~\cite{cp2017, cp2019, cp2023}.

\noindent \textbf{1) Filter pruning.}
Most pruning approaches rely on the "smaller-norm-less-important" criterion, which involves pruning filters with lower norm values in the network \cite{li2017pruning, ye2018rethinking}. However, He et al.~\cite{FPGM} point out the limitations of this criterion-based approach. They propose a novel technique for calculating the Geometric Median of filters within the same layer. By doing so, they prune filters that make the most replaceable contribution instead of those with comparatively less contribution. Criterion-based pruning methods tend to reduce model capacity due to fixed pruning thresholds. To address this, He et al.~\cite{LFPC} introduce learnable pruning thresholds for each layer using a differentiable criterion sampler, which can be updated during training. Additionally, Zhang et al.~\cite{ASTER} propose an adaptive pruning threshold based on the sensitivity of the loss to the threshold value.

\noindent \textbf{2) Channel pruning.} 
Channel pruning is another effective approach for reducing FLOPs and inference time, complementing filter pruning. He et al.~\cite{cp2017} first implement channel pruning by focusing on eliminating redundant channels by evaluating the L1 norm. Peng et al.~\cite{cp2019} take a different approach by using the Hessian matrix to model inter-channel dependencies and select channels using sequential quadratic programming. For more complex modules like group convolutions and depthwise convolutions, Liu et al.~\cite{cp2021} introduce a layer grouping mechanism to search for coupled channels automatically. The importance of these channels is calculated based on Fisher's information. CATRO~\cite{cp2023} leverages feature space discrimination to assess the joint impact of multiple channels while consolidating the layer-by-layer impact of preserved channels.

\subsubsection{Comparison of pruning methods}
Table~\ref{tab:pruningMethods} displays the accuracy after pruning and the corresponding pruned FLOPs of the various structure pruning methods. While one might initially assume that the best-performing methods prune the highest number of FLOPs, in reality, we often perceive the "best" as those that effectively balance the trade-off between pruned FLOPs and the associated drop in accuracy. For instance, while GFP attains the highest pruned accuracy, its reduction in FLOPs is limited to 50.6\%. In contrast, ASTER removes the most FLOPs, yet its pruned accuracy falls short of being the best. In summary, filter and channel pruning methods can efficiently decrease the FLOPs while maintaining similar accuracy. We advocate choosing a pruning method that seamlessly integrates with the current network architecture, prioritizing ease of implementation. For example, if the network's feature map boasts over a thousand channels but only uses a few filters, opting for channel pruning would be more beneficial.

\begin{table}[t]
\centering
\footnotesize
\caption{Comparison of different pruning methods using ResNet50 on the ImageNet dataset. The methods that achieve the highest percentage of pruned FLOPs are marked in \textbf{bold}.
}
\vspace{-0.5em}
\begin{tabular}{|cc|ccc|}
\hline
\textbf{Type} & \textbf{Method (30\%)} & \textbf{Baseline (\%)} & \textbf{Pruned Acc. (\%)} & \textbf{Pruned FLOPs (\%)} \\ 
\hline
- & ResNet50 & 76.15 & - & -  \\ \hline
\multirow{4}{*}{Filter}
 & SFP~\cite{SFP} & 76.15 & 74.61 (-1.54) & 41.8 \\
 & FPGM~\cite{FPGM}  & 76.15 & 75.59 (-0.56) & 42.2  \\ 
 & LFPC~\cite{LFPC} & 76.15 & 74.46 (-1.69) & 60.8  \\ 
 & ASTER~\cite{ASTER} & 76.15 & 75.27 (-0.88) & \textbf{63.2}  \\ \hline
\multirow{4}{*}{Channel} & CCP~\cite{cp2019} & 76.15 & 75.50 (-0.65) & 48.8  \\ 
 & GFP~\cite{cp2021} & 76.79 & \textbf{76.42 (-0.37)} & 50.6\\ 
 & SCP~\cite{cp2020} & 75.89 & 74.20 (-1.69) & 54.3  \\
 & CATRO~\cite{cp2023} & 75.98 & 75.84 (-0.14) & 45.8  \\ \hline
\end{tabular}
\label{tab:pruningMethods}
\vspace{-1em}
\end{table}

\begin{figure*}[b] 
\centering 
    		\includegraphics[width=0.9\textwidth]{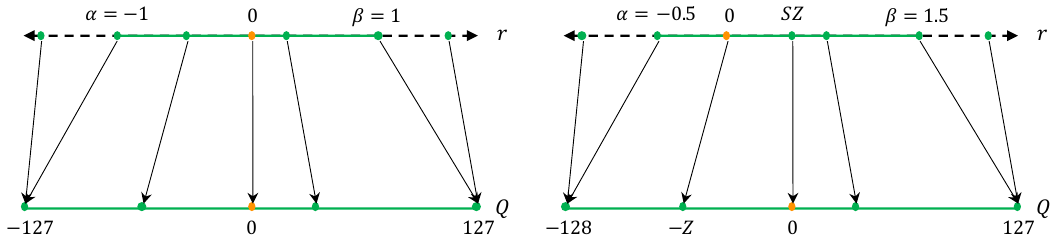}
                \vspace{-1em}
    		\caption{Symmetric (left) and asymmetric (right) quantization representation~\cite{gholami2022survey}. Note that r represents the real value, S represents the real-valued scaling factor, and Z represents the integer zero point.} 
    		\label{fig:quantTypes}
\end{figure*}

\subsection{Quantization}
Pruning is an efficient way to compress the model. However, after pruning, the remaining weights, typically stored as full-precision 32-bit floating-point numbers (float32), still demand significant memory. To address this, quantization~\cite{gray1998quantization}, a technique that allows parameters to be represented with reduced bit precision, becomes a desirable solution. Specifically, quantization maps weights and activations to a set of finite numbers through a calibration process that determines potential values using a symmetric or asymmetric representation. As depicted in Fig.~\ref{fig:quantTypes}, both methods define a range [$\alpha$, $\beta$], in symmetric quantization, -$\alpha$ = $\beta$, whereas in the asymmetric quantization, −$\alpha \neq \beta$.  

The calibration of this range, as outlined by Gholami et al.~\cite{gholami2022survey}, falls into two categories: dynamic and static calibration. The first one is accurate but computationally demanding, as it computes $[\alpha, \beta]$ for each feature map. The latter is a computationally lighter alternative because it calculates the range based on typical values after several iterations, albeit with less accuracy. Both dynamic and static calibration are pivotal for optimizing the quantization process.

Quantization theory has been applied to NN from various perspectives over time. For instance, Gupta et al.~\cite{DBLP:journals/corr/GuptaAGN15} introduce the use of fixed-point numbers during the model's training process to enhance the algorithm's noise tolerance. They also employ stochastic rounding as an alternative to the round-to-nearest strategy to counteract the adverse effects of fixed-point numbers. In another approach, Faghri et al.~\cite{Adaptive_Quant_2020} introduce two adaptive quantization methods, Adaptive Level Quantization (ALQ) and Adaptive Multiplier Quantization (AMQ), which update their compression method in parallel during training to quantize the gradients in data-parallel stochastic gradient descent adaptively. This adaptation aims to reduce communication costs between the processors. Lastly, Wang et al.~\cite{Wang_2022_CVPR} treat the quantization problem as a differentiable lookup operation. They jointly optimized both the network and the associated tables during training.

\subsubsection{Half-precision and Mixed-precision training}
Mixed-precision training involves using lower-precision values while retaining full-precision values for crucial information~\cite{Mixed_PT}. For instance, in a notable series of works, HAWQ~\cite{dong2019hawq} implements an automatic approach based on the Hessian of the model to determine the optimal mixed-precision settings for weight values. Subsequently, the HAWQ-V2 model~\cite{hawqv2} introduces mixed-precision quantization for activation values. The HAWQ-V3 model~\cite{yao2021hawq} further improves it by focusing on integer-only quantization. Interestingly, Liu et al.~\cite{liu2021post} introduce a method that utilizes a linear combination of multiple low-bit vectors to approximate a full-precision vector. This approach achieves "mixed-precision training" with a single precision level by varying the number of vectors to approximate different weights.


\subsubsection{Quantization using fewer bits}
In an early work by Banner et al.~\cite{2018_8bit}, the quantization of weights, activations, and most gradient streams in all layers of an NN is performed using 8-bit precision by replacing traditional batch-norm with ranged batch-norm layers. Another technique proposed by Wang et al.~\cite{wang2018training} allows matrix and convolutional operations to also be implemented using 8-bit numbers. Furthermore, there have also been methods that use ternary values to quantize an NN. In an important work done by Liu et al., TWN~\cite{TWN_2023} manages to constrain weights to +1, 0, and -1 values, achieving a 16x compression of the model. This idea is extended in TTQ~\cite{TTQ_2017}, where the positive and negative weights use two different learnable scales $w_{1}$ and $w_{2}$, resulting in possible values of $-w_{1}$, $0$, and $w_{2}$.

More aggressive approaches have sought to reduce quantization levels further by implementing NN binarization. This approach uses binary values instead of floating-point or integer values for faster computations, lower memory usage, and reduced power consumption. Courbariaux et al.'s pioneering work~\cite{courbariaux2016binarized} binarizes networks by restricting the weights and activations to either +1 or -1, determining the final values by evaluating the sign of the real values. Variations of this work include topologies such as XNOR-Net~\cite{rastegari2016xnor} and the Least Squares method~\cite{pouransari2020least}, which introduce an additional activation layer after the binary convolutions.

\subsubsection{Quantization Aware Training (QAT)}
In the early stages of quantization research, a prevalent approach was first to train an unquantized model, apply a quantization process, and then retrain or fine-tune the model to achieve an acceptable level of accuracy. This methodology, known as Post-training Quantization (PTQ), proved to be an effective strategy for achieving significant compression, especially when the pre-trained model has ample representational capacity. The success of PTQ lies in its ability to balance compression gains and maintain satisfactory model accuracy, making it a pivotal technique in model optimization and deployment. However, quantization is a lossy process, which can lead to a significant drop in model accuracy. To address this issue, Jacob et al.~\cite{Quan_int2018} introduced QAT, a technique that computes inference-time quantization errors during the model training stage, allowing the model to become aware of these errors and make adjustments accordingly. This process simulates inference-time errors through a process known as FakeQuant.


Improvements to the core QAT technique have been explored by introducing learnable clipping scalars~\cite{choi2018pact}.
In a recent development, Sakr et al.~\cite{sakr2022optimal} achieved state-of-the-art performance by identifying the MSE-minimizing clipping scalars and implementing 4-bit quantization.

\subsubsection{Comparison of quantization methods}
Table~\ref{tab:quantMethods} compares the performance of quantization methods on the ImageNet dataset, emphasizing the trade-off between compression and accuracy loss. Notably, binarized networks aiming for a 32x compression and speedup show significant accuracy drops. On the other hand, approaches with 4-bit quantization, except~\cite{liu2021post}, result in little loss of accuracy and can, therefore, be a good choice of precision for quantization. 
However, theoretical compression and speedup expectations may not align with actual results due to additional operations like quantization and dequantization. This may explain why some works opt not to conduct an in-depth analysis of the quantized model size, although~\cite{liu2021post} does provide such an analysis and successfully achieves an approximately 8-fold reduction in the model size (42.56 MB to 5.37 MB). 

\begin{table}[b]
    \centering
    \footnotesize
    \vspace{-1em}
    \caption{Comparison of several quantization methods using different levels of precision to compress a ResNet18 on the ImageNet dataset.}
    \vspace{-1em}
    \label{tab:quantMethods}
    \begin{tabular}{|c|ccc|}
        \hline
        \textbf{Method} & \textbf{Initial Accuracy}. (\%) & \textbf{Quantized accuracy} (\%) & \textbf{Precision} \\
        \hline
        QIL~\cite{jung2019learning} & 70.2 & 70.1 (-0.1) & 4-bit \\
        \cite{liu2021post} & 69.8 & 61.7 (-8.1) & 4-bit \\
        LLT~\cite{Wang_2022_CVPR} & 69.8 & 70.4 (+0.6) & 4-bit \\
        LLT~\cite{Wang_2022_CVPR} & 69.8 & 69.5 (-0.3) & 3-bit \\
        HAWQ-V3~\cite{yao2021hawq} & 71.5 & 68.5 (-3.0) & MP \\
        TWN~\cite{TWN_2023} & 65.4 & 61.8 (-3.6) & 2-bit \\
        TTQ~\cite{TTQ_2017} & 69.6 & 66.6 (-3.0) & 2-bit \\
        XNOR-Net~\cite{rastegari2016xnor} & 69.3 & 51.2 (-18.1) & 1-bit \\
        Least Squares~\cite{pouransari2020least} & 69.6 & 63.4 (-6.2) & 1-bit \\
        \hline
    \end{tabular}
    \vspace{-1.3em}
\end{table}

\subsection{Knowledge Distillation (KD)}
\begin{figure}[b] 
\centering 
            \includegraphics[width=\columnwidth]{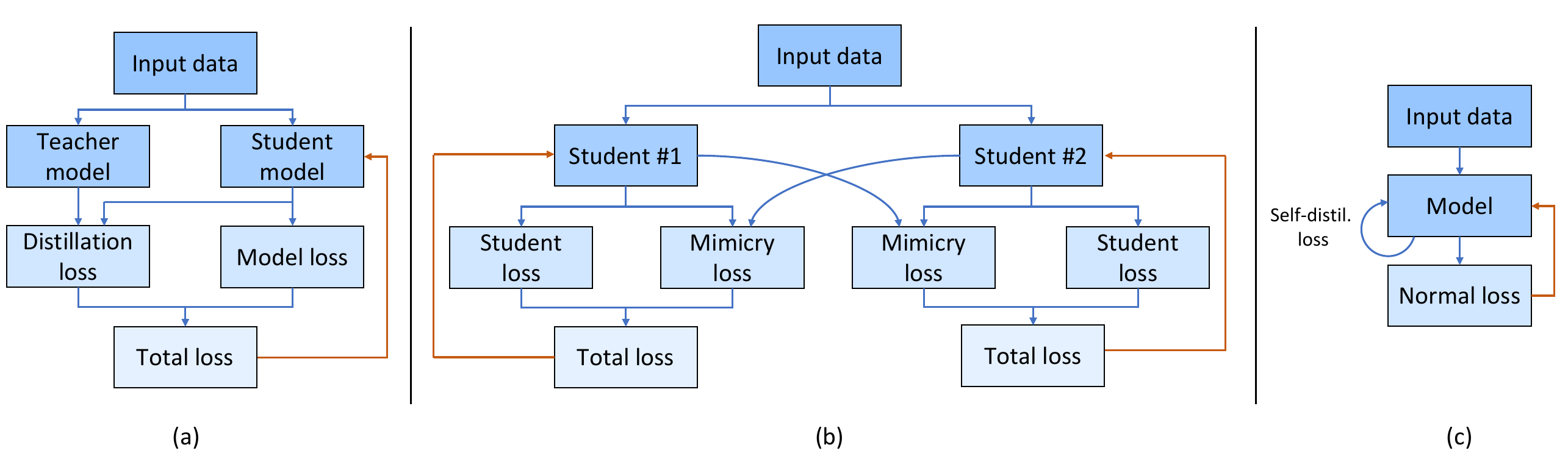}
    	\vspace{-2em}	
            \caption{(a) Offline Distillation~\cite{hinton2015distilling}. (b) Online Distillation~\cite{zhang2017deep}. (c) Self-Distillation~\cite{zhang2019your}. We use orange lines to indicate the gradient update.} 
    	\label{fig:KD_final}
    \vspace{-1em}
\end{figure}

KD is a model compression technique designed to transfer knowledge from a large network to a smaller one~\cite{hinton2015distilling, gou2021knowledge}. Its simplest form is illustrated in Fig.~\ref{fig:KD_final}(a), where the larger model is referred to as the teacher and the smaller model as the student.
In the approach proposed by Hinton et al.~\cite{hinton2015distilling}, the teacher model is initially trained to generate soft labels. Then, the training of the student model leverages ground-truth labels and the teacher's predictions on the same data. This combination enables the student to attain performance comparable to the teacher using fewer parameters.

KD algorithms can be categorized into three types: offline, online, and self-distillation, as illustrated in Fig.~\ref{fig:KD_final}. The key distinction lies in the teacher's definition and training strategy. For instance, in offline distillation, teacher and student training processes are performed sequentially, whereas in online distillation, the teacher can continue or initiate training alongside the student. On the other hand, in self-distillation, the student becomes its own teacher.
\vspace{-0.3em}

\subsubsection{Offline Distillation}
Most of the earlier KD works fall under the category of offline distillation. In offline distillation, a pre-trained teacher model is required, as seen in the case of the vanilla KD~\cite{hinton2015distilling}. While offline distillation is relatively easy to implement, it comes with the unavoidable overhead of time and computational resources required to train a large teacher model first.

Various methods have been explored to enhance KD algorithms, including introducing alternative loss functions such as contrastive-based loss~\cite{contrastive_2020} and minimizing the maximum mean discrepancy between models~\cite{huang2017like}. Significant size disparities between teacher and student models can impact results, leading Zhao et al.~\cite{Zhao_2022_CVPR} to redefine logit distillation by decoupling the influence of target and non-target classes. Lin et al.~\cite{Lin_2022_CVPR} address the semantic information gap in KD by dynamically distilling each pixel of the teacher features to all spatial locations of the student features, guided by a similarity measure from the transformer.

Recently, SimKD~\cite{chen2022knowledge} proposed a straightforward distillation approach, reusing the teacher's classifier and aligning intermediate features with an L2 loss. SemCKD~\cite{chen2021cross} involves student learning through feature embedding, preserving feature similarities in the intermediate layers of the teacher network.

\subsubsection{Online Distillation}
Offline distillation can be problematic when obtaining a pre-trained large teacher model is not feasible, rendering many of the previously mentioned methods unusable. Online distillation introduces an end-to-end training strategy that overcomes this limitation by concurrently training the teacher and student networks, challenging the traditional concept of a "single large teacher"~\cite{zhang2017deep, guo2020online, li2022distilling}.

The Deep Mutual Learning (DML) algorithm, proposed in~\cite{zhang2017deep}, eliminates the need for a pre-trained teacher in the KD process, as depicted in Fig.~\ref{fig:KD_final}(b). Instead, this approach advocates simultaneous learning of a cohort of networks, with each network incorporating the predictions of the others in its loss functions. This change enables all networks in the cohort to benefit from each other's knowledge, even improving networks that are large enough to have acted as teachers in a conventional KD process. These large networks can enhance their results with knowledge distilled from other untrained, smaller networks.
Further refinements of this approach have been made in~\cite{guo2020online,li2022distilling}. Online distillation techniques can also incorporate adversarial concepts. Zhang et al.~\cite{zhang2021adversarial} propose an adversarial co-distillation approach that employs Generative Adversarial Networks (GANs) to explore "divergent examples" and enhance knowledge transfer.

Furthermore, online distillation has demonstrated notable efficacy in scenarios requiring generating pseudo labels for data. The widely adopted mean teacher framework~\cite{tarvainen2017mean} introduces the concept of employing two identical models; specifically, the teacher model has the same structure as the student model. The primary idea involves updating the teacher's weights through an exponential moving average (EMA) of the student's weights. In various unsupervised contexts \cite{deng2021unbiased, yu2022cross}, this principle is leveraged to create pseudo labels for training the student via a supervised loss. Notably, each prediction made by the teacher model can be viewed as an ensemble incorporating the current and past iterations of the student model, rendering it inherently more robust and stable.

\subsubsection{Self-Distillation}
As depicted in Fig.~\ref{fig:KD_final}(c), self-distillation techniques involve the process of KD, where a model distills knowledge from itself. In this scenario, during the training process, a single instance of the model simultaneously acts as both the teacher and student. Strategies in this distillation approach encompass using the same model saved at different epochs~\cite{yang2019snapshot} and leveraging various model layers for self-instruction~\cite{yuan2021revisiting, hou2019learning}.

Zhang et al.~\cite{zhang2019your} pioneered self-distillation from deeper to shallower layers of the model. Their innovation improves results and reduces training time by eliminating the need for additional networks. Similarly, Hou et al.~\cite{hou2019learning} harness knowledge transfer through attention maps from deeper layers. Yang et al.~\cite{yang2019snapshot} use the weights of previous iterations for knowledge distillation instead of using deeper layers of the model. Kim et al.~\cite{kim2021self} elevate self-distillation with a sophisticated progressive framework, incorporating adaptive gradient rescaling for hard example mining. 

In an important study, Yuan et al.~\cite{yuan2021revisiting} challenge the foundations of conventional KD by introducing the Teacher-free KD (Tf-KD). They explore the intricate relationship between KD and Label Smoothing Regularization (LSR) techniques and suggest employing self-training or manually-designed regularization terms for improving the student model's accuracy when faced with the difficulty of a powerful teacher model. 
Additionally, self-distillation methods have successfully been applied to domain adaptation tasks~\cite{yoon2022semi, sultana2022self}.


\begin{table}[b]
\vspace{-1em}
\caption{KD methods evaluated on the CIFAR-100 dataset. ↑ indicates an improvement over the baseline. Note: The pair of accuracies in the online distillation methods represent the teacher and student models' performances after distillation.}

\label{tab:kdMethods}
\footnotesize
\centering
\vspace{-0.5em}
\begin{tabular}{|l|l|l|l|l|} 
\hline
\textbf{Methodology}  & \textbf{Algorithm} & \textbf{Teacher (baseline)} & \textbf{Student (baseline)} & \textbf{Improved Accuracy} \\ 
\hline
 \multirow{4}{*}{Offline distillation} &  SimKD~\cite{chen2022knowledge} & ResNet32 (79.42)    & ResNet8 (73.09)  & 78.08 (4.99 ↑)         \\
 & SemCKD~\cite{chen2021cross}                     & ResNet32 (79.42)            & ResNet8 (73.09)       & 76.23 (3.14 ↑)         \\
 & SRRL~\cite{yang2021knowledge}               & ResNet32 (79.42)               & ResNet8 (73.09)        & 75.39 (2.30 ↑)         \\
 & SemCKD \cite{chen2021cross} & ResNet32 (79.42)         & WRN-40-2 (76.35)               & 79.29 (2.94 ↑)         \\

\hline
  \multirow{4}{*}{Online distillation} & DML \cite{zhang2017deep}    & WRN-28-10 (78.69)           & WRN-28-10 (78.69)           & 80.28, 80.08 (1.39 ↑)  \\
 & DML \cite{zhang2017deep}    & WRN-28-10 (78.69)           & ResNet32 (68.99)            & 78.96, 70.73 (1.74 ↑)  \\
 & FFSD \cite{li2022distilling}    & ResNet56 (71.55)           & ResNet32 (69.96)           & 75.78, 74.85 (4.90 ↑)  \\
 & KDCL \cite{guo2020online}                       & WRN-16-2 (72.20)            & ResNet32 (69.90)            & 75.50, 74.30 (4.40 ↑)  \\
\hline
 \multirow{4}{*}{Self-distillation} & SD \cite{yang2019snapshot}                      & –                           & ResNet32 (68.39)            & 71.29 (2.90↑)          \\
 & Tf-KD \cite{yuan2021revisiting}                 & –                           & ResNet18 (75.87)            & 77.10 (1.23↑)          \\
 & PS-KD \cite{kim2021self}                & –                           & ResNet18 (75.82)            & 79.18 (3.36↑)          \\
 & Tf-KD \cite{yuan2021revisiting}                 & –                           & ShuffleNetV2 (70.34)        & 72.23 (1.89↑)          \\
\hline
\end{tabular}
\end{table}

\subsubsection{Comparison of KD methods}
Table~\ref{tab:kdMethods} compares several distillation methods and analyzes their respective outcomes on the CIFAR-100 dataset. These findings challenge the perception that offline distillation methods are outdated and too simplistic. For example, SimKD recently achieved state-of-the-art performance with a ResNet32 as the teacher and a ResNet8 as the student. 
Additionally, our analysis demonstrates the efficacy of online distillation, showcasing instances where a teacher can improve its own performance despite instructing a student with significantly lower accuracy. Notably, the WRN-28-10 achieves a 0.27\% (78.69\% to 78.96\%) improvement even when paired with a ResNet32 that initially achieves nearly 10\% (78.69\% to 68.99\%) less accuracy.
Furthermore, self-distillation emerges as a promising strategy, necessitating only one model, exemplified by a ResNet18 achieving 3.36\% gains through the PS-KD method, albeit not surpassing the improvements seen in other methods. To address this limitation, it is advisable to complement self-distillation with other forms of distillation or compression methods for enhanced performance.
Ultimately, a comparison between methodologies is hard, as performance heavily depends on implementation details. Therefore, we advocate for adopting a strategy that is easier to implement and aligns most logically with the ongoing development objectives.

\subsection{Neural Architecture Search (NAS)}
Even if DL techniques excel in numerous tasks, it is true that they often depend heavily on human expertise to find the best trade-off between performance and complexity. Optimizing a model can be exceptionally challenging due to a multitude of choices involving hyperparameters, network layers, hardware devices, etc.

In response to this challenge, Automated Machine Learning (AutoML), which aims to automatically build ML systems without much requirement for ML expertise and human intervention, is being extensively studied~\cite{he2021automl}. Several mature tools exist for AutoML applications, such as Auto-WEKA~\cite{kotthoff2019auto} and Auto-sklearn~\cite{feurer2019auto}. In this paper, our primary focus is NAS, a crucial section of AutoML. 
The fundamental concepts of NAS are outlined as follows:
\begin{itemize}
    \item \textbf{Search Space:} The search space encompasses the possible combinations of hyperparameters, including kernel size, channel size, convolution stride, depth, and more. A larger search space that covers a wider range of possibilities increases the likelihood of discovering a highly accurate model. However, a vast search space can lead to longer search times.

    \item \textbf{Search Algorithm:} This refers to the algorithm used to find the optimal combination within the search space. Common strategies include random search, grid search, reinforcement learning (RL)~\cite{zoph2016neural,tan2019mnasnet}, evolutionary algorithms (EA)~\cite{enas,xue2023neural}, and gradient optimization~\cite{liu2018darts,wu2019fbnet}. An efficient search strategy can significantly reduce search time, especially in extensive search spaces.

    \item \textbf{Performance Evaluation Strategy:} This defines the criteria for selecting the neural architecture that maximizes specific performance metrics among all the models generated through NAS. Performance metrics, such as Top-1 or Top-5 scores for classification and average precision (AP) or F1 scores for object detection, reflect the suitability of the hyperparameter combinations for the given task.
\end{itemize}

In this section, we explore various approaches in the field of NAS, including RL-based NAS, EA-based NAS, Gradient-based NAS, and other related works, all based on different search algorithms.

\subsubsection{RL-based NAS}
\begin{figure}[b] 
\centering 
    		 \includegraphics[width=0.6\columnwidth]{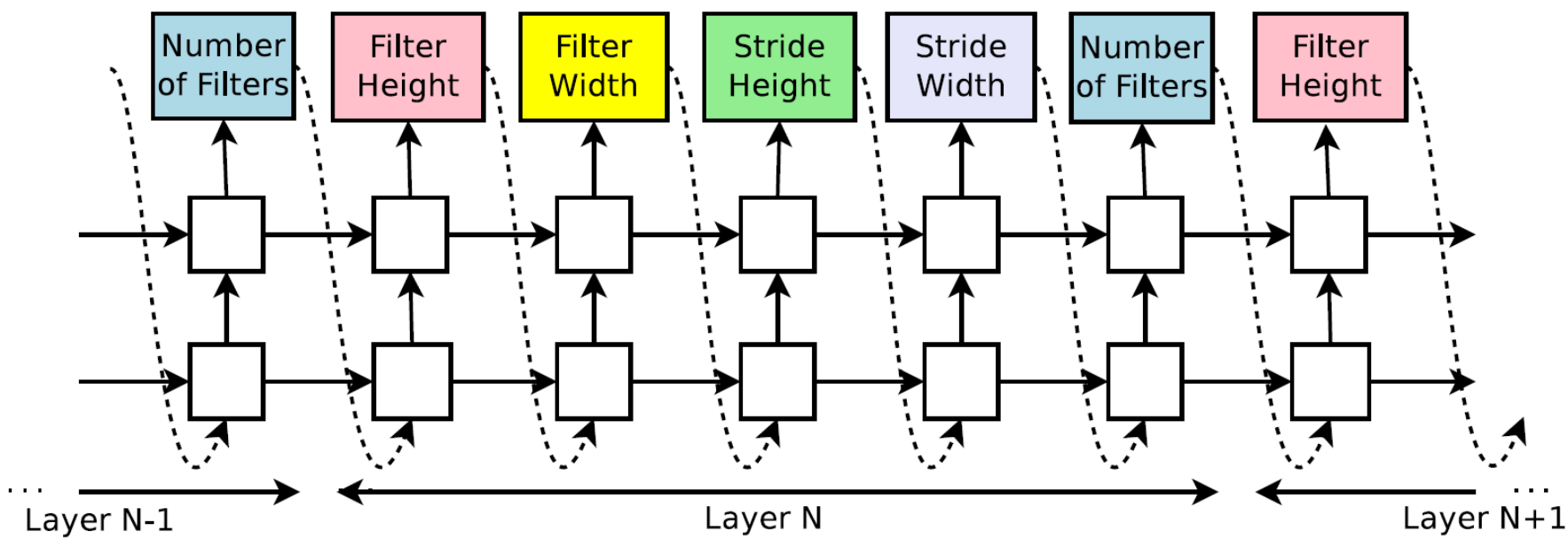}
                \vspace{-1em}
    		\caption{NAS with RL~\cite{zoph2016neural}.} 
                \vspace{-1.3em}
    		\label{fig:NASRL}
\end{figure}

In this pioneering work of adopting RL for NAS, Zoph et al.~\cite{zoph2016neural} utilize a recurrent neural network (RNN) controller (called an agent) to generate candidate hyperparameters for constructing child networks (environments). The child network then receives a score (reward) based on metrics like accuracy and AP. The RNN controller updates itself according to the reward and refines the hyperparameters for the child network iteratively. A detailed process is illustrated in Fig.~\ref{fig:NASRL}. Moving forward, MnasNet~\cite{tan2019mnasnet} considers latency and employs RL to identify Pareto optimal solutions that balance latency and performance. This approach also introduces a factorized hierarchical search space, which organizes the CNN into predefined blocks and explores different connections and operations within each block.

\vspace{-0.2em}
\subsubsection{EA-based NAS}
To enhance model performance, Real et al.~\cite{enas} introduce an EA-based approach for NAS. This method continuously evolves model architectures. The evolution process begins with workers generating an initial set of models, forming what is known as a population. During the evolution step, two models are randomly selected from the population, and their accuracy on the validation set is evaluated. The weaker-performing model is removed from the population, while the better model becomes the parent model. In the mutation step, the parent model is duplicated, producing two identical copies. One of these copies is reintroduced into the population, while the other undergoes mutation to create a new model, referred to as the child model. Subsequently, the workers train and assess the child model's performance before adding it back to the population. This process is iteratively repeated, resulting in increasingly improved models within the population.

However, a random search approach within a large population can be highly inefficient when dealing with a vast search space. To address this concern, Sun et al.~\cite{enas3} develop an encoding mechanism that maps CNN features to numerical values. This enables the acceleration of the evolutionary process by using a CNN architecture as an input to the Random Forest. More recently, Xue et al.~\cite{xue2023neural} proposed a queue mechanism to reduce the population and incorporate crossover and mutation operators to enhance the diversity of child networks.

\subsubsection{Gradient-based NAS}
The core concept of gradient-based NAS involves the transformation of a discrete search space into a continuous one, enabling the application of gradient descent techniques to discover optimal model architectures automatically. Inferring latency after each training is inefficient for the proposed NAS network, especially for research institutes with limited resources. Additionally, using gradient-based NAS methods is deemed more appropriate when formulating hardware-aware NAS approaches.

{\centering
\begin{figure}[b] 
	\includegraphics[width=0.65\columnwidth]{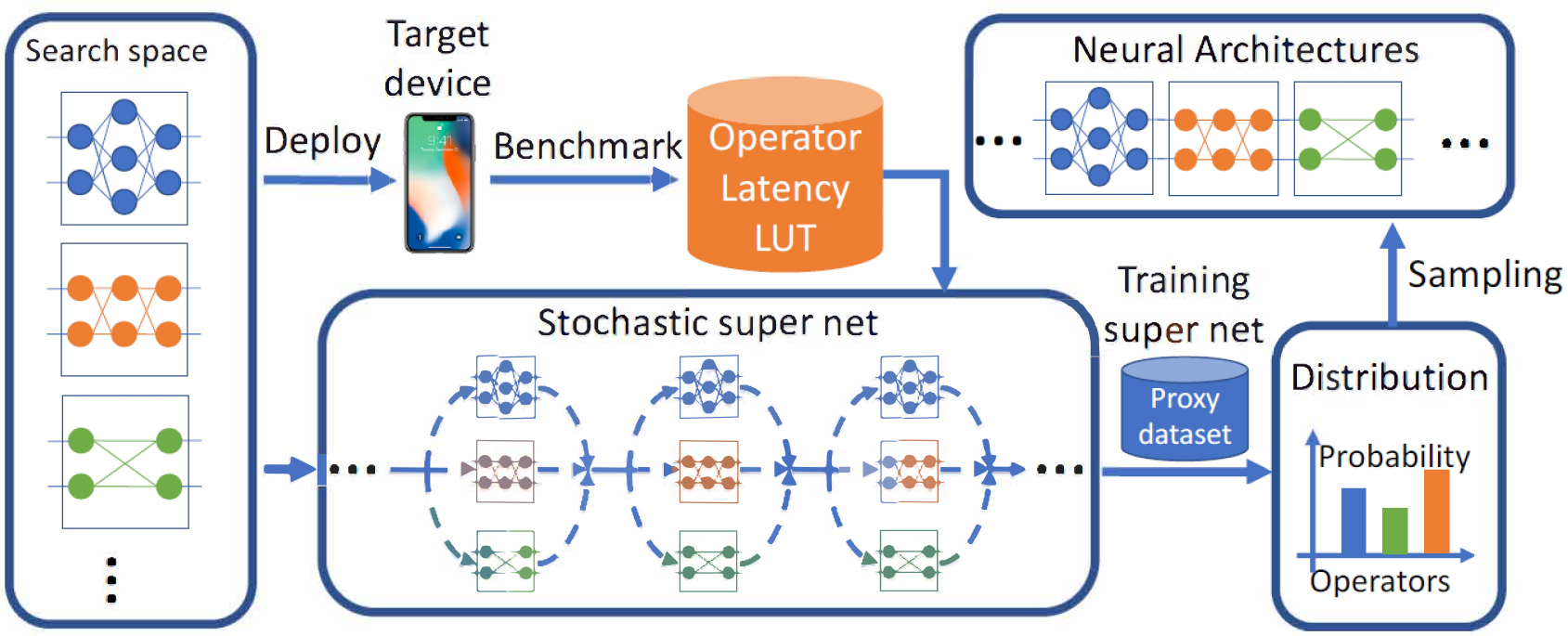}
	\caption{The DNAS pipeline in FBNet~\cite{wu2019fbnet}.} 
	\label{fig:DNAS}
\end{figure}
}

DARTS~\cite{liu2018darts} presents an efficient architecture search algorithm based on gradient descent that avoids black-box search problems. It converts structural parameters from discrete to continuous, making them differentiable. As a result, DARTS provides accurate, efficient, and differentiable NAS.
Inspired by works such as MnasNet~\cite{tan2019mnasnet}, DARTS~\cite{liu2018darts}, and NetAdaptV1~\cite{yang2018netadapt}, FBNet~\cite{wu2019fbnet} is a hardware-aware NAS breakthrough discovered through the differentiable NAS (DNAS) pipeline, depicted in Fig.~\ref{fig:DNAS}. In FBNet, nine distinct blocks are designed within the layer, and 22 layers are utilized to construct a stochastic supernetwork, which is optimized using stochastic gradient descent (SGD). Additionally, FBNet devises a layer-wise search space, enabling each layer to select a different block. Furthermore, in order to reduce the layer-wise search space with lower latency, a latency lookup table is employed, and a latency-aware loss term is incorporated into the overall loss function, given by:
\begin{equation}
    L(a, w_a) = CE(a, w_a) \cdot \alpha \log (LAT(a))^\beta.
    \label{eq:1}
\end{equation}
where $a$ and $w_a$ denote the network architecture and network parameters for a specific device, while $CE$ represents the cross-entropy loss. $LAT$ stands for the latency of the architecture on the target device, which is determined using a lookup table.
The parameters $\alpha$ and $\beta$ serve as the magnitude of the overall loss function and the latency term, respectively. 
For further details and related work on FBNet, please refer to \cite{wan2020fbnetv2,dai2021fbnetv3}.

\subsubsection{Other NAS related works}
Numerous other NAS algorithms have been proposed. One example is the Symbolic DNN-Tuner~\cite{2022symbolic}, which introduces an automatic software system for determining optimal tuning actions following each network training session using probabilistic symbolic rules. The system comprises a module for data processing, search space exploration, and Bayesian optimization. The controller module manages the training process and decides the tuning actions.
Besides finding the best combination from a vast search space, testing the proposed combination network is also time-consuming. Measuring the latency of the entire model on the target device each time can be highly inefficient.

\begin{figure}[t] 
\centering 
	\includegraphics[width=0.5\columnwidth]{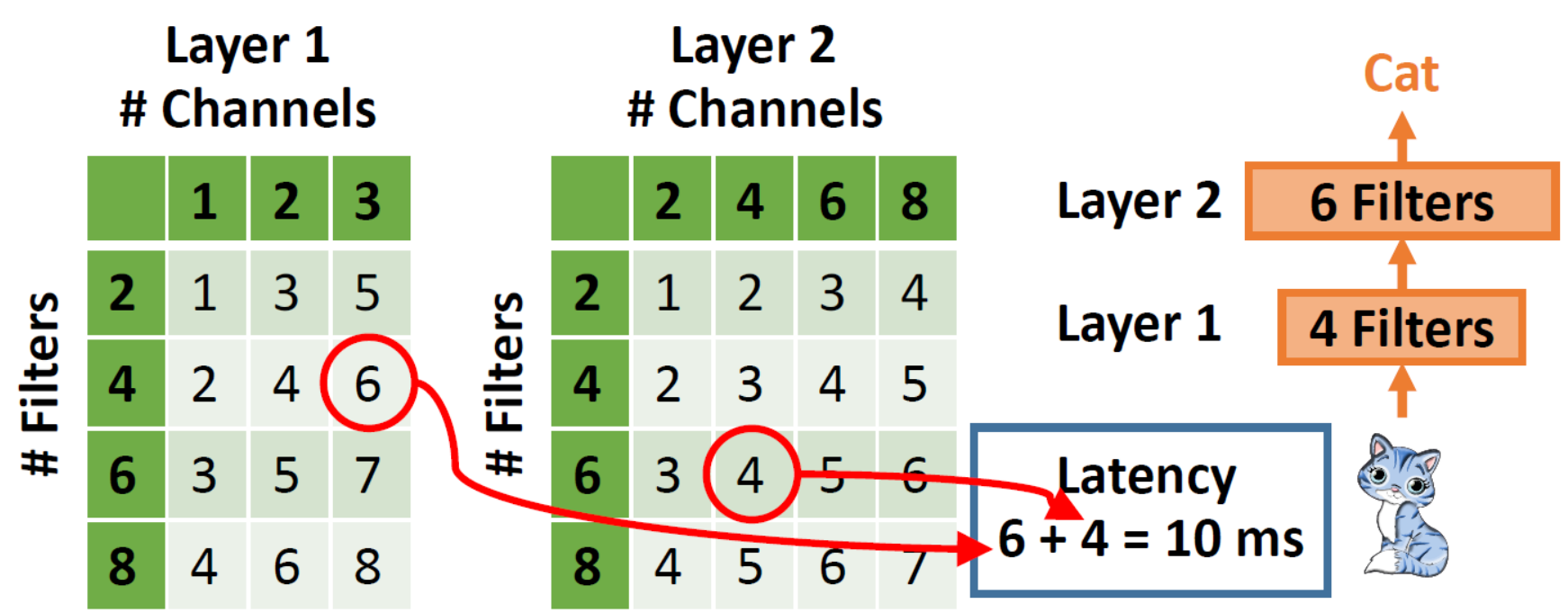}
        \vspace{-1em}
	\caption{Layer-wise look up table~\cite{yang2018netadapt}.} 
	\label{fig:Layer-wise look up table}
   \vspace{-1em}
\end{figure}

To address this issue, NetAdaptV1~\cite{yang2018netadapt} employs an adaptive algorithm that considers energy consumption and memory usage, enabling it to respond more realistically to hardware constraints. The approach involves the creation of a layer-wise lookup table, as shown in Fig.~\ref{fig:Layer-wise look up table}, simplifying the search complexity for a pre-trained network.
In this setup, the latency of each layer is pre-measured, and a lookup table is constructed to record latency based on the layer's structure. For instance, as illustrated in Fig.~\ref{fig:Layer-wise look up table}, Layer 1 consists of 3 channels with 4 filters and a measured latency of 6 ms, and Layer 2 consists of 4 channels with 6 filters and a measured latency of 4 ms. The total latency is calculated as the sum of the latency for each layer, resulting in a total latency of 10 ms (6 + 4).

Moving forward, NetAdaptV2~\cite{yang2021netadaptv2} introduces Channel-Level Bypass Connections (CBCs), which combine depth and layer width in the original search space to enhance the efficiency of both training and testing.
Moreover, Abdelfattah et al.~\cite{abdelfattah2021zero} leverages pruning-at-initialization~\cite{lee2018snip} and incorporates six zero-cost proxies for NAS proposal scoring. This innovative approach requires only a single minibatch of data and a single forward/backward propagation pass instead of full training, resulting in a more efficient NAS process.

\subsection{Discussion and Summary}
This section encapsulates a summary of the preceding discussion on model compression. Additionally, it provides valuable practical tips and guidance, aiming to offer actionable insights for effective implementation and application in relevant contexts.

\noindent \textbf{Pruning.} Although unstructured pruning methods~\cite{obd,lottery} have made significant strides in parameter reduction, their irregular structures frequently pose compatibility issues with hardware accelerators. Therefore, structure pruning~\cite{FPGM,SFP,cp2023} has emerged as a preferable alternative, primarily due to its regular structure. Notably, modern DL frameworks, such as PyTorch and TensorFlow, have integrated built-in functionalities that facilitate the seamless implementation of structure pruning. This streamlined integration enhances the ease and efficiency with which structure pruning techniques can be applied.

\noindent \textbf{Quantization.} When considering quantization, the choice of technique depends on the hardware environment where the model will be deployed. Hardware specifications play a critical role, turning quantization from an optional optimization into an imperative requirement. For instance, specific MCUs or edge TPUs exclusively support integer operations, making full integer quantization essential for model implementation. TensorFlow Lite (TF-Lite)~\cite{Tensorflow_qa}  effectively addresses this need, reducing the model size by up to four times and significantly accelerating inference by more than three times. In hardware with low-power CPUs, an 8-bit integer quantization strategy is often recommended, as CPUs exhibit exceptional computational efficiency when handling integer operations instead of floating-point values. Notably, when using 16-bit float quantization, values are subsequently de-quantized back to 32-bit float representations during execution on the CPU. For a deeper analysis of hardware support for quantization and facilitating libraries, see~\cite{liang2021pruning}.

\noindent \textbf{Knowledge distillation.} KD techniques have significantly enhanced NNs by leveraging insights from other models. In practice, the offline KD process \cite{hinton2015distilling} can be effectively utilized when training a large model is viable. On the other hand, online distillation stands forth as a promising solution. For example, the DML process \cite{zhang2017deep} has shown remarkable results without necessitating a pre-trained teacher model, making it adaptable to multi-GPU training with several small models. In situations characterized by a scarcity of labeled data or noisy labels, the mean teacher framework has emerged as a valuable and effective solution. Moreover, self-distillation and ongoing advancements in KD \cite{Lin_2022_CVPR, Zhao_2022_CVPR} open numerous possibilities for exploration and offer different options for the definition of the teacher and student networks.

\noindent \textbf{NAS.} While both RL-based NAS \cite{tan2019mnasnet} and EA-based NAS~\cite{enas3} have demonstrated their capacity to achieve impressive accuracy, it is important to note that their training demands extensive resources and time, often spanning days or weeks and involving hundreds of GPUs. This resource-intensive nature has contributed to a relative decline in the number of studies in these areas. Therefore, when confronted with GPU limitations, gradient-based algorithms like DARTS~\cite{liu2018darts} and FBNet~\cite{wu2019fbnet}, which introduce continuity into the search space, can be considered. This approach significantly reduces the training time. Alternative options include approaches like "once for all" NAS~\cite{cai2019once}, which tailor the extensive network into subnetworks optimized for different target devices. However, if ample computational resources are at hand, RL-based and EA-based NAS methods are viable options, and they also offer superior performance compared to gradient-based NAS~\cite{ren2021comprehensive}. Additionally, when memory footprint, energy consumption, and latency are key considerations, the hardware-aware NAS concepts introduced by studies like FBNet~\cite{wu2019fbnet}, NetAdapt~\cite{yang2018netadapt}, and NetAdaptV2~\cite{yang2021netadaptv2} may be particularly relevant.

\noindent \textbf{Conclusion.} In conclusion, model compression approaches have their strengths and limitations. Quantization is a relatively simple but proven effective compression technique in many cases. It is essential to first match the selected quantization approach with the specific hardware requirements for floating-point or integer values. In scenarios where hardware constraints permit, starting with a 16-bit float quantization is often a prudent initial step. If there is a need for more substantial model compression, two viable options emerge. First, model pruning offers an effective solution, substantially reducing redundant network parameters while preserving performance integrity. This is particularly valuable when working with resource-constrained environments. Secondly, the KD framework proves advantageous, especially in scenarios with ample unlabeled data, as often encountered in applications like autonomous driving. The mean teacher structure, in particular, is a valuable tool for generating pseudo labels from unlabeled data, effectively incorporating this additional information into training and enhancing overall model performance. Finally, NAS can also be considered, particularly for tasks where it excels the most, such as image classification, where it can potentially discover optimal network architectures tailored to specific requirements. The choice among these approaches should be guided by the specific demands of the task and the available computational resources.

\section{Hardware Acceleration of Deep Learning Models}
\label{section:CNN Accelerator for Hardware system} 
With the advancements in GPUs, DL has risen to the forefront of artificial intelligence technology. DL models, such as CNNs, are computationally intensive. Hence, hardware acceleration is becoming imperative to render DL applications feasible and practical. In this section, we present an overview of prominent hardware accelerators of DL models. We then introduce typical dataflow and data locality optimization techniques, as well as widely adopted DL libraries. Finally, we discuss algorithms that employ a co-design approach for software/hardware deployment.

\subsection{Hardware Architectures} 
Hardware accelerators for DL models encompass a range of options, including GPUs and CPUs based on temporal architecture, as well as FPGAs and ASICs rooted in spatial architecture. The basic components of a hardware accelerator are an arithmetic logic unit (ALU), a control unit, and a local memory unit (cache unit). In the temporal architecture, the control and local memory units are centralized, and the processing elements (PEs) only contain the ALUs. Data is accessed sequentially from centralized memory to PEs, with no interactions between the PEs~\cite{tem_spa}. In contrast, spatial architecture entails PEs equipped with control units, ALUs, and local memory (register file). This allows independent data processing and direct communication between PEs.  

\subsubsection{Temporal Architecture} 
Temporal architectures are often adopted in general-purpose platforms, like CPUs and GPUs, which are optimized for sequential tasks and parallel tasks, respectively. 

\noindent \textbf{Central processing unit (CPU).}
CPUs process input data into usable information output, executing calculations sequentially through serial computing. A recent CPU-based acceleration technique, SLIDE~\cite{cpu1}, which leverages C++ OpenMP to combine intelligent randomized algorithms with multi-core parallelism and workload optimization, demonstrates that employing smart algorithms on a CPU can potentially achieve better speed than using an NVIDIA-V100 GPU. 
 
\noindent \textbf{Graphics processing unit (GPU).}
GPUs are designed for parallel computation. Their architecture may consist of thousands of cores. Hence, GPUs excel at parallel computing, enabling them to process multiple instructions simultaneously, making them highly efficient for tasks that involve simple and repetitive computations. Given that DL models often entail extensive matrix addition and multiplication operations, GPUs have emerged as the primary accelerators for the development of DL. Their parallel processing capabilities make them instrumental in accelerating DL tasks.

\subsubsection{Spatial Architecture}  
By utilizing PEs, spatial architectures often seen in FPGAs and application-specific integrated circuits (ASICs), the necessity for repeated and redundant access to external memory is reduced, leading to lower energy consumption.

\noindent \textbf{FPGAs.} FGPAs consist of programmable logic blocks with logic gates capable of performing computations. Reprogrammable by nature, they can accelerate various DL structures effectively and better support pruning methods. 
Additionally, FPGAs can directly implement algorithms without any decoding and interpretation process. To enhance AI applications using FPGAs, Qi et al.~\cite{fpga3} emphasize key concepts of parallel computing and demonstrate how these concepts can be implemented in FPGAs. 
Roggen et al.~\cite{fpga4} successfully implement digital signal processing (DSP) algorithms, such as filter finite impulse response filters on FPGA platforms, thereby improving support for wearable computing. 
For more references on FPGA AI applications, consult~\cite{2023fpga,embedded21}. 

\noindent \textbf{ASICs.}
ASICs, customized for specific electronic systems, outperform FPGAs with superior speed, lower power consumption, and higher throughput. TPUs, prominent ASICs tailored for AI applications~\cite{tpu1}, excel in efficiently executing matrix operations, a pivotal capability advantageous in deep learning computations with prevalent expansive matrix multiplications. In a recent development, the newly introduced TPU-v3 can connect 1024 TPU chips through a 2-D torus network~\cite{tpu3}. This innovation enhances parallelism and enables execution on more TPU-v3 accelerator cores through spatial partitioning and weight update-sharing mechanisms. The supercomputer TPU-v4~\cite{tpu4} further elevates the capabilities by increasing the number of TPU chips to 4096. TPU-v4 also introduces optical circuit switches (OCSes) that dynamically restructure their interconnection topology to improve scalability, accessibility, and utilization. As a result, TPU-v4 offers a 2.7 times improvement in performance/watt and a tenfold increase in speed compared to TPU-v3.

\subsubsection{Discussion of CNN Accelerators} 
CPUs are generally not well-suited for training and inference of typical DL models due to low FLOPs performance.
GPUs, which can support parallel computation with thousands of cores, excel in parallel computing and are widely adopted in various AI applications. However, GPUs are known for their high power consumption, rendering them unsuitable for edge devices and IoT applications. On the other hand, FPGAs and ASICs offer more energy-efficient acceleration options for edge AI applications. The choice between FPGAs and ASICs often depends on the specific requirements.
FPGAs are preferred for AI products that require rapid development or are produced in small batches. ASICs are more suitable for AI products that undergo mass production, especially highly mature or customized ones. For projects with ample budget, TPUs can be the top choice. TPUs boast exceptional computational power, making them ideal for handling extensive models with large batch sizes, such as the GPT-4~\cite{gpt4} and LLaMA~\cite{llama}, significantly reducing training and inference times.

\subsection{Dataflow and the Data Locality Optimization}
The computational complexity and data storage demands of CNNs pose significant challenges to computational performance and energy efficiency. These challenges are particularly pronounced in smaller devices with limited memory, including constrained on-chip buffers (SRAM) and off-chip memory (DRAM). To address these issues, optimizing dataflow is crucial for enhancing memory and energy efficiency. The dataflow process in deep models generally consists of three main steps. Firstly, DL models are stored in off-chip memory, often referred to as external memory. Secondly, when convolution kernels are required, they are fetched from on-chip buffers. Finally, PEs are employed to execute the MACs.

\subsubsection{Dataflow types} 
Hardware accelerators of DL models have different types of dataflow based on their applications and can be categorized into pipeline-like dataflow~\cite{li2016multistage,lin2017data}, DaDianNao-like dataflow~\cite{luo2016dadiannao,chen2014dadiannao}, Systolic-array-like dataflow~\cite{tpu1,wei2017automated,zhang2018caffeine}, and streaming-like dataflow~\cite{du2017reconfigurable,guo2017angel}. 

\noindent \textbf{Pipeline-like dataflow.}
In this dataflow, the input pixels (the pixels of the feature map) are passed on to individual PEs, and the model's weights (representing model parameters) are fixed on each PE. Notably, the partial sum is then forwarded to the subsequent PE. This approach offers substantial parallelism, facilitating the concurrent processing of data by multiple stages, thereby enhancing computational efficiency. However, tasks are executed sequentially, with each stage dependent on the completion of the previous one, potentially resulting in increased latency.

\noindent \textbf{DaDianNao-like dataflow.}
In this dataflow, each PE can function like a neuron, processing input pixels in a way akin to an NN. Specifically, input pixels are routed to each PE, and the model's weights are embedded within each PE. The computed partial sums are then aggregated using an adder tree. This type of dataflow can 
accommodate different kernel sizes, making it capable of handling intricate and irregular model structures. However, 
this dataflow approach is energy-intensive and demands substantial hardware resources due to the model's complexity.

\noindent \textbf{Systolic-array-like dataflow.}
This dataflow sequentially conveys input pixels and weights into the PEs, with PEs cascaded to enhance computational efficiency. Subsequently, an adder tree is employed to aggregate the partial sums. This dataflow approach optimizes the utilization of hardware resources, improves overall hardware efficiency, and mitigates timing issues in large designs. However, finding an appropriate mapping for CNNs onto a systolic array can be challenging.

\begin{figure}[b]
\centering 
\includegraphics[width=1\columnwidth]{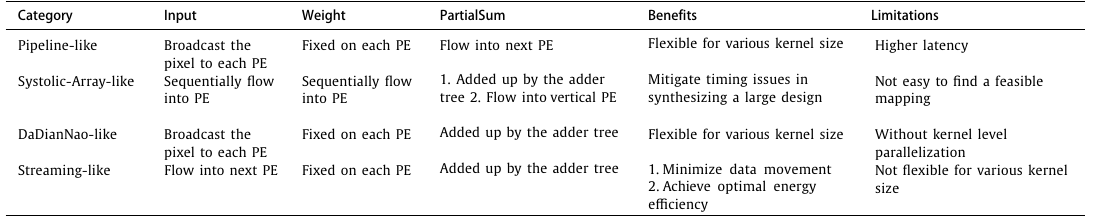}
\vspace{-2em}
\caption{A comparison of dataflow types~\cite{hsu2020essa}. PE stands for processing element.}
\vspace{-0.8em}
\label{fig:compare}
\end{figure}

\noindent \textbf{Streaming-like dataflow.}
In this dataflow, input pixels are continuously sent to the following PE without pausing or needing intermediate storage, with weights being fixed on each PE. Subsequently, the adder tree accumulates the partial sums. This dataflow is particularly suitable for streaming data, such as audio and video processing, due to its high throughput and low latency. Nonetheless, applications requiring complex operations between stages or that rely on previous results may require additional processing and design. Fig.~\ref{fig:compare} compares the types of dataflow. 
\vspace{-0.5em}

\subsubsection{Data Locality Optimization}
CNNs deliver exceptional performance characterized by high throughput and energy consumption. However, their performance can be restricted by limited on-chip memory. Therefore, an effective locality optimization mechanism is essential. Data locality optimization focuses on devising a dataflow schedule that maximizes data reuse utilization and minimizes data movement. A prevalent approach involves applying loop transformation techniques, such as loop unrolling, loop tiling, and loop interchange, to optimize NN deployment. These techniques help maximize hardware utilization and minimize memory traffic, addressing the limitations of on-chip memory constraints.

Loop unrolling~\cite{unrolling1,unrolling2} is a method that involves expanding loop iterations into multiple sequential instructions.
This technique significantly reduces the number of loop iterations in the CNN, resulting in faster CNN operations and improved hardware utilization through increased parallelization. However, it is important to note that loop unrolling may lead to code bloat, increased memory usage, and higher storage requirements, especially for larger CNN models.

Loop tilling~\cite{qiu2016going,looptiling,looptiling3} involves partitioning the input data into several blocks to enable parallel computations for CNN acceleration. For example, an original input data of size $224 \times 224 \times 3$ can be divided into smaller blocks of size $112 \times 112 \times 3$. These smaller blocks are processed sequentially to mitigate buffer loading and memory constraints. This technique effectively adapts to limited on-chip memory and significantly enhances cache locality. However, for modern accelerators, such as GPUs, where memory access patterns are already optimized for high throughput, loop tilling may add extra complexity without appreciable gains in performance. 

Loop interchange~\cite{loopinter,loopinter-2} involves changing the order of loops within a nested loop with the aim of improving data locality and extracting parallelism. Specifically, the order of the loops is optimized to allow each iteration of the outermost loop to utilize the same cache line, hence reducing memory access. Loop interchange can also accelerate CNN models by increasing the use of operators like addition and multiplication. Notably, some algorithms have complex intrinsic properties and special meanings in their loop orders. Therefore, altering the loop order may yield meaningless results and reduce performance.

In this section, we introduce typical types of dataflow and provide an overview of various mechanisms for data locality optimization. 
More in-depth details can be found in 
~\cite{fu2023optimizing,wolf1991data}.

\subsection{Deep Learning Libraries}
To facilitate the deployment of a DL model, it is also essential to use DL libraries that provide high-level APIs to simplify the implementation, design, and training of complex NNs. We introduce several popular DL libraries supporting GPU acceleration and the auto gradient system.

TensorFlow~\cite{abadi2016tensorflow} supports static and dynamic graphs, allowing users to select the most suitable mode. With this flexibility, TensorFlow supports the research and development of custom DL models. Additionally, TensorFlow also provides extensive APIs for DL model implementation. For instance, a TensorFlow model can be converted into a TensorFlow-Lite (TF-Lite)~\cite{TFlite} model, a smaller, more efficient ML model format that can be run on mobile and edge devices. 

PyTorch~\cite{paszke2019pytorch} is a framework renowned for its remarkable capacity to facilitate the creation of intricate models and the fine-tuning of NNs down to the minute details, making it a favored choice within the research community. Its simplicity, user-friendliness, and intuitiveness made it a go-to tool for prototyping DL models. However, there are certain deployment-related limitations with its API, which might restrict its application in certain real-life scenarios.
 
MXNet~\cite{chen2015mxnet} is a library that provides optimized building blocks for implementing CNNs. It is specially tailored for Intel processors, offering vectorized and threaded support for CNNs on Intel CPUs and GPUs. Moreover, the MXNet framework provides interfaces in multiple languages, including Python, Scala, Java, Clojure, and R, making it convenient for cross-domain DL developers.
 
NVIDIA has been at the forefront of GPU hardware and software optimization for DL. cuDNN~\cite{chetlur2014cudnn} is a highly optimized library specifically designed for DL networks, providing acceleration for DNN-related tasks. In addition to cuDNN, NVIDIA offers a range of DL libraries included in CUDA-X~\cite{nvidiaCUDAX}. TensorRT~\cite{vanholder2016efficient}, another NVIDIA library, optimizes inference on NVIDIA GPUs by applying layer and tensor fusion, kernel auto-tuning, and dynamic tensor memory optimizations.

Each DL library has unique strengths and caters to specific use cases, allowing practitioners to choose one that best suits their projects. To address the interoperability challenges between DL libraries, Microsoft and Facebook introduced Open Neural Network Exchange (ONNX)~\cite{onnx}, an open standard for machine learning interoperability. With ONNX, models created in different libraries can be easily shared and executed. For instance, a PyTorch model can be run on an Android device by converting it into TensorFlow format, eliminating the need for model retraining. 

\subsection{Co-Design of Hardware Architecture}
In DL, acceleration solutions relying solely on software techniques are primarily limited by their dependence on the intrinsic capabilities of general-purpose processors, potentially struggling to exploit specialized hardware features designed for specific DL tasks fully. Conversely, hardware-only solutions may face limitations in flexibility and adaptability, as dedicated hardware is often tailored for specific tasks or architectures, making updates or adaptations to new DL models challenging without hardware modifications. This underscores the value of co-designing a hardware and software approach for resource-constrained environments, employing a holistic optimization strategy. 
This approach includes refining the DL algorithm, optimizing and compressing the model, efficient memory management, software kernel implementation, and hardware architecture design. 
This section discusses solutions that adopt a holistic approach to address challenges related to irregular memory accesses, enhance the handling of sparsity resulting from compression methods, and explore improved solutions within NAS algorithms.

In Section 3, we emphasize that many NN connections can be pruned effectively without substantial accuracy loss. However, in such models, only a subset of the NN's weights are active, and their locations are irregular or non-contiguous. Efficiently accessing these weights, especially when using hardware accelerators like GPUs or TPUs, can be challenging due to the irregularity of weight locations. To tackle this issue, in earlier methods, like Cambricon-X~\cite{zhang2016cambricon}, MAC operations utilize zero-weight connections and access required weights using sparse indices. However, irregular nonzero weight distribution caused issues such as indexing overhead, PE imbalances, and inefficient memory access. Later advancements, as seen in Cambricon-S~\cite{zhou2018cambricon}, improve efficiency by enforcing regularity in filter sparsity through software/hardware integration. 

Sparse-YOLO~\cite{wang2020sparse} introduces a dedicated sparse convolution unit tailored to handle quantized values and sparsity resulting from unstructured pruning techniques. Cho et al.~\cite{cho2021reconfigurable} propose an acceleration technique for a quantized binary NN. This approach utilizes an array of PEs, with each PE responsible for computing the output of a specific feature map, implementing inter-feature map parallelism. Moreover, optimizing the storage of sparse weights post-pruning has been explored. Han et al.~\cite{han2015deep} show that these sparse weights can be compressed, reducing memory access bandwidth by around 20\%-30\%. SCNN~\cite{parashar2017scnn} processes convolutional layers in their compressed format using an input stationary dataflow. This involves transmitting compressed weights and activations to a multiplier array, followed by a scatter network to add the scattered partial sums.

In the NAS field, apart from the previously discussed hardware-aware NAS approaches that tailor models for specific hardware platforms, there are also co-designed solutions that initially remain hardware-agnostic. These co-designed systems seamlessly integrate hardware optimization within the NAS process, ensuring simultaneous hardware and DNN model optimization. Hardware settings can be explored in conjunction with DNN architectures using the same algorithm~\cite{zhou2021rethinking, choi2021dance, li2020edd} or through an external search algorithm~\cite{sekanina2021neural, lin2019neural}.

\begin{figure}[t]
\centering 
\includegraphics[width=0.85\columnwidth]{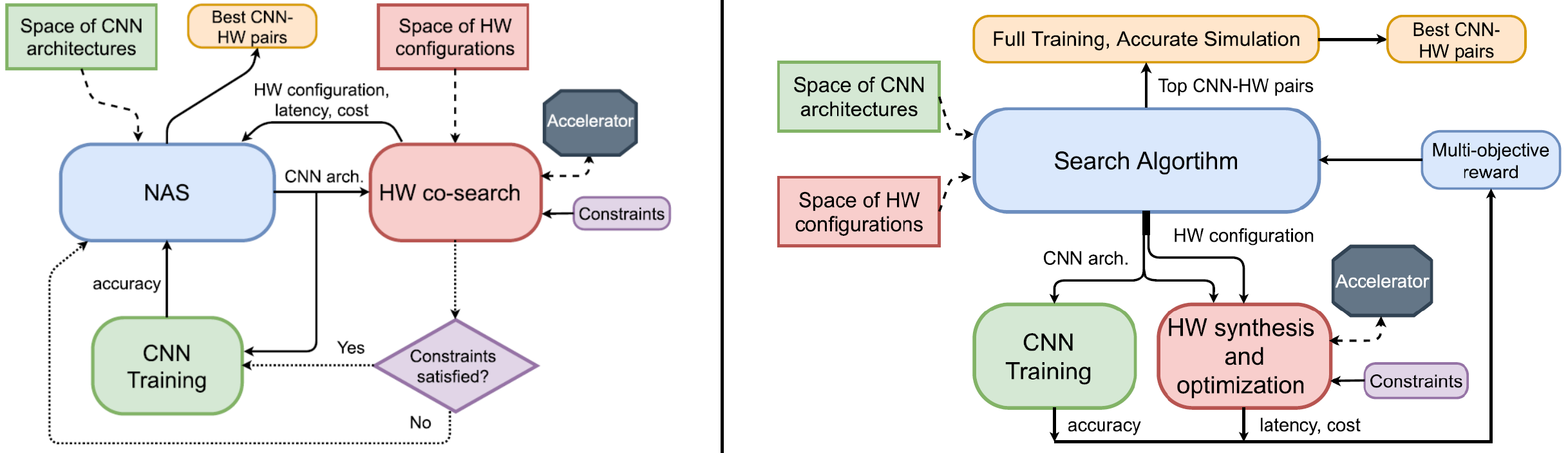}
\caption{Two different approaches for implementing NAS and hardware co-design~\cite{sekanina2021neural}.}
\label{fig:codesignNAS}
\vspace{-2em}
\end{figure}

As shown in Fig.~\ref{fig:codesignNAS}(a), the most direct approach for co-searching hardware and software settings involves creating CNN and accelerator pairs and evaluating the final model's performance. One can opt to train the CNN each time a new pair is tested or follow the approach of Chen et al. \cite{chen2020you}, where a supernet is employed to directly generate the weights of a DDN, and accuracy is assessed in a single testing run of the model. 
Fig.~\ref{fig:codesignNAS}(b) illustrates an alternative strategy employed by Lin et al.~\cite{lin2019neural}, 
where a hardware optimization algorithm takes a candidate CNN as input and optimizes the hardware accelerator to achieve specific objectives. The network is then trained and evaluated only if a viable hardware configuration is found. If no suitable hardware setting is identified, the network remains untrained until a viable configuration is found. This strategy allows for the avoidance of training the CNN, which is the most complex phase of the co-design process.


In summary, the co-design of algorithms significantly improves compression and computational efficiency. However, these methods are inherently non-trivial and require in-depth exploration of software and hardware techniques. 

    
\section{Challenge and Future work}
\label{section:Challenge and Future work}
In this survey, we explore the sophisticated domain of lightweight models, compression methods, and hardware acceleration, showcasing their advanced technological capabilities applicable across a broad spectrum of general applications. Nonetheless, deploying these models in resource-constrained environments continues to present substantial challenges. This section is dedicated to unveiling novel techniques in TinyML and LLMs for accelerating and applying DL models, focusing on unresolved issues that warrant further investigation.
\vspace{-1em}

\subsection{TinyML}
TinyML is an emerging technology that enables DL algorithms to run on ultra-low-end IoT devices that consume less than 1mW of power. However, the extremely constrained hardware environment makes it challenging to design and develop a TinyML model. Low-end IoT devices predominantly employ MCUs due to their cost efficiency compared to CPUs and GPUs. However, MCU libraries, such as CMSIS-NN~\cite{lai2018cmsis} and TinyEngine~\cite{lin2020mcunet}, are often platform-dependent, unlike GPU libraries like PyTorch and TensorFlow, which offer cross-platform support. Consequently, the design focus of TinyML leans more toward specialized applications rather than facilitating general-purpose research, potentially impeding the pace of overall research advancements.

\noindent \textbf{MCU-based libraries.}
Due to the resource-constrained environments in TinyML, MCU-based libraries are often designed for specific use cases. 
For instance, CMSIS-NN~\cite{lai2018cmsis}, a pioneering work for MCU-based libraries developed on ARM Cortex-M devices, proposes an efficient kernel divided into NNfunctions and NNsupportfunctions. NNfunctions execute the main functions in the network, such as convolutions, poolings, and activations. NNsupportfunctions contain data conversions and activation tables. CMIX-NN \cite{capotondi2020cmix} proposes an open-source mixed and low-precision tool that can support the model's weights and activation to be quantized into 8, 4, and 2 bits arbitrarily. 
MCUNet~\cite{lin2020mcunet} presents a co-design framework tailored for DL implementation on commercially available MCUs. This framework incorporates TinyNAS to search for the most accurate and lightweight model efficiently. Additionally, it leverages the TinyEngine, which encompasses code generator-based compilations and in-place depthwise convolution, effectively addressing peak memory constraints. Moving forward, MCUNetV2~\cite{mcunetv2}  introduces a patch-based inference mechanism that operates only on a small spatial region of the feature map, further reducing peak memory use. MicroNet~\cite{banbury2021micronets} adopts differentiable NAS (DNAS) to search for efficient models with a low number of operations and supports the open-source platform Tensorflow Lite Micro (TFLM). MicroNet achieves state-of-the-art results for all TinyMLperf industry-standard benchmark tasks, i.e., Visual Wake Words, Google Speech Commands, and Anomaly detection.

\noindent \textbf{What hinders the rapid development of TinyML?}
Despite its progress, the growth of TinyML is hindered by several inherent key constraints, including resource constraints, hardware and software heterogeneity, and lack of datasets~\cite{ray2022review}. Extreme resource constraints, such as an incredibly small size of SRAM and less than 1 MB size flash memory, pose challenges in designing and deploying TinyML models on edge devices. Furthermore, due to hardware heterogeneity and a lack of framework compatibility, current TinyML solutions are tweaked for every individual device, complicating the wide-scale deployment of TinyML algorithms. Besides, existing datasets may not be suitable for TinyML architecture as the data may not correspond to the data generation feature from external sensors of edge devices. A set of standard datasets suitable for training TinyML models is needed to advance the development of effective TinyML systems. These open research challenges need to be addressed before mass deployment on IoT and edge devices is possible.

\subsection{Building lightweight Large Language Models}
LLMs have consistently exhibited outstanding performance across various tasks in the past two years~\cite{anil2023palm, touvron2023llama, ye2023comprehensive}. LLMs hold significant potential for practical applications, especially when paired with human supervision. For instance, they can serve as co-pilots alongside autonomous agents or as sources of inspiration and suggestions. However, 
these models typically feature parameters at the billion scale. Deploying such models for inference generally demands GPU-level hardware and tens of gigabytes of memory, posing substantial challenges for everyday LLM utilization. 
For example, Tao et al.~\cite{tao2022compression} find it hard to quantize generative pre-trained language models due to homogeneous word embedding and varied weight distribution. Consequently, transforming a large, resource-intensive LLM model into a compact version suitable for deployment on resource-constrained mobile devices has emerged as a prominent future research direction.

World-renowned enterprises have made significant strides in LLM deployment. In 2023, Qualcomm showcased the independent execution of the text-to-image model, Stable Diffusion~\cite{stablediff} and the image-to-image model, ControlNet~\cite{controlnet} on mobile devices, thereby accelerating the deployment of large models to edge computing environments. Google also introduced several versions of its latest universal large model, PaLM 2~\cite{anil2023palm}, featuring a lightweight variant tailored for mobile platforms. This development has created new opportunities for migrating large models from cloud-based systems to edge devices. 
However, certain large models still require several gigabytes of physical storage and runtime memory. Consequently, efforts are being directed towards achieving a memory footprint of less than 1 GB~\cite{ray2022review}, signifying that significant work is still needed in this area. This section outlines some key initiatives for easing the implementation of LLMs in resource-constrained environments.

\vspace{-0.5em}
\subsubsection{Pruning without re-training}
Recently, a substantial body of work has applied common DL quantization and pruning techniques to construct lightweight LLMs. Some approaches~\cite{yu2023boost, wu2023understanding} focus on implementing quantization, where numerical precision is greatly reduced.  
SparseGPT~\cite{frantar-sparsegpt} demonstrates, for the first time, that large-scale Generative Pre-trained Transformer (GPT) models can be pruned to at least 50\% sparsity in a single step, without any subsequent retraining, with minimal loss of accuracy. Following this, Wanda (Pruning by Weights and Activations)~\cite{wanda}, specifically designed to induce sparsity in pre-trained LLMs, is introduced. Wanda prunes weights with the smallest magnitudes and does not require retraining or weight updates. The pruned LLM can be directly utilized, increasing its practicality. Notably, Wanda surpasses the established baseline of magnitude pruning and competes effectively with recent methods that involve extensive weight updates. These works set a significant milestone for future work in designing LLM pruning methods that do not require retraining. 

\vspace{-0.3em}
\subsubsection{Model Design}
From a model design perspective, we can create lightweight LLMs from the very inception, focusing on reducing the number of model parameters. One promising avenue in this endeavor is prompt tuning, which seeks to optimize the LLMs' performance while maintaining efficiency and model size.
A notable approach in this context is Visual Prompt Tuning (VPT)~\cite{jia2022visual}, which emerges as an efficient and effective alternative to the comprehensive fine-tuning of large-scale Transformer models employed in vision-related tasks. VPT introduces a mere fraction, less than 1\%, of trainable parameters within the input space while maintaining the integrity of the model's backbone. Another noteworthy contribution is CALIP~\cite{guo2023calip}, which introduces parameter-free attention mechanisms to facilitate effective interaction and communication between visual and text features. It yields text-aware image features and visual-guided text features, contributing to the development of more streamlined and efficient vision-language models. In the near future, one promising avenue for advancing lightweight LLM design is the development of adaptive fine-tuning strategies. These strategies would dynamically adjust the model's architecture and parameters to align with specific task requirements. This adaptability ensures the model can optimize its performance for particular applications without incurring unnecessary parameter bloat. 

\subsubsection{Building Lightweight Diffusion Model}
In recent years, denoising diffusion-based generative models, particularly those of the score-based variety~\cite{ho2020denoising, song2020score}, have made notable strides in creating diverse and authentic data. However, the transition of the inference phase of a diffusion model to edge devices poses significant challenges. The inference phase reverses the transformation process to generate real data from Gaussian noise, commonly known as the denoising process.
Moreover, when these models are compressed to reduce their footprint and computational demands, there is a potential risk of severe degradation in image quality. The compression process may need simplifications, approximations, or even the removal of essential model components, which could adversely affect the model's ability to reconstruct data from Gaussian noise accurately. Consequently, a critical concern emerges in balancing model size reduction with preserving high-quality image generation, thereby presenting a formidable challenge in developing diffusion models in resource-constrained scenarios. 

In a very recent work, Shang et al.~\cite{shang2023post} introduce post-training quantization~\cite{cai2020zeroq} into the field of diffusion model acceleration. When applied in a training-free manner, this quantization approach exhibits the capability to enhance the efficiency of the denoising process while simultaneously reducing the storage requirements for diffusion model weights, a critical component in the acceleration of diffusion models. Nevertheless, there remain numerous opportunities for improvement in this domain to achieve a trade-off between high-quality and lightweight model solutions.
\subsubsection{Deployment of Vision Transformers (ViTs)}
Despite the increasing prevalence of lightweight ViTs, deploying ViT in hardware-constrained environments remains a persistent concern. According to~\cite{wang2022towards}, ViT inference on mobile devices has a latency and energy consumption of up to 40 times higher than CNN models. 
Hence, without modification, mobile devices cannot support the inference of ViTs. 
The self-attention operations in ViTs need to compute the pair-wise relations between image patches, and the computations grow quadratically with the number of patches. 
Moreover, computation for FFN layers is more time-consuming than attention layers~\cite{wang2022towards}. By removing the redundant attention heads and FFN layers, DeiT-Tiny can reduce latency by 23.2\%, with negligible 0.75\% accuracy loss.

Several works designed NLP models for embedded systems such as FPGAs~\cite{ham20203, ham2021elsa, wang2021spatten}. 
More recently, DiVIT~\cite{li2022divit} and VAQF~\cite{sun2022vaqf} proposed hardware-software co-designed solutions for ViTs. 
DiVIT proposes a delta patch encoding and novel differential attention at the algorithm level that leverages the patch locality during inference. 
In DiVIT, the design of a differential attention Processing Engine array with bit-saving techniques can calculate the delta with less computation and communicate with differential dataflow. Furthermore, the exponent operation is executed using a lookup table without additional computation and with minimal hardware overhead.
VAQF first introduces binarization into ViTs, which can be used for FPGA mapping and quantization training. Specifically, VAQF can generate the required quantization precision and accelerator description for direct software and hardware implementation based on the target frame rate.

To enable the seamless deployment of ViTs in resource-constrained devices, we highlight two potential future directions:

\noindent \textbf{1) Algorithm optimizations.} 
In addition to the design of efficient ViT models described in Section 2.3, the bottlenecks of ViTs should also be considered. For example, since MatMul operations cause a bottleneck in ViTs, these operations can be accelerated or reduced~\cite{wang2022towards}. Additionally, integer quantization and improvement to operator fusion can be considered.

\noindent \textbf{2) Hardware Accessibility.}
Unlike CNNs, which are well-supported on most mobile devices and AI accelerators, ViTs do not have specialized hardware support. For instance, ViT fails to run on mobile GPUs and Intel NCS2 VPU.
Based on our findings, some important operators are not supported on specific hardware. Specifically, on the mobile GPU, the concatenate operator requires a 4-dimensional input tensor in TFLiteGPUDelegate, but the tensor in ViTs is 3-dimensional. On the other hand, Intel VPU does not support LayerNorm, which exists in the architecture of transformers but is uncommon in CNN. Hence, hardware support for ViTs on resource-constrained devices warrants further investigation.

\section{Conclusion}
\label{Conclusion}

Recently, computer vision applications have increasingly prioritized energy conservation, carbon footprint reduction, and cost-effectiveness, highlighting the growing importance of lightweight models, particularly in the context of edge AI. This paper conducts a comprehensive examination of lightweight deep learning (DL), exploring prominent models such as MobileNet and Efficient transformer variants, along with prevalent strategies for optimizing these models, including pruning, quantization, knowledge distillation, and neural architecture search. Beyond providing a detailed explanation of these methods, we offer practical guidance for crafting customized lightweight models, offering clarity through an analysis of their respective strengths and weaknesses.

Furthermore, we discussed hardware acceleration for DL models, delving into hardware architectures, distinct data flow types and data locality optimization techniques, and DL libraries to enhance comprehension of accelerating the training and inference processes. This investigation sheds light on the intricate interplay between hardware and software (Co-design), providing insights into expediting training and inference processes from a hardware perspective. Finally, we turn our gaze toward the future, recognizing that the deployment of lightweight DL models in TinyML and LLM technologies presents challenges that demand the exploration of creative solutions in these evolving fields.

\section{Acknowledgement}
This work is partially supported by the National Science and Technology Council, Taiwan under Grants, NSTC-112-2628-E-002-033-MY4, NSTC-112-2634-F-002-002-MBK, and NSTC-112-2218-E-A49-023, and was financially supported in part (project number: 112UA10019) by the Co-creation Platform of the Industry Academia Innovation School, NYCU, under the framework of the National Key Fields Industry-University Cooperation and Skilled Personnel Training Act, from the Ministry of Education (MOE) and industry partners in Taiwan.

\bibliographystyle{ACM-Reference-Format}
\bibliography{main}

\end{document}